\documentclass[pdflatex,sn-mathphys-num]{sn-jnl}

\usepackage{parskip}        
\usepackage{amsmath, amssymb}  
\usepackage{graphicx}       
\usepackage{float}          
\usepackage{caption}        
\usepackage{subfigure}      
\usepackage{rotating}       
\usepackage{pdflscape}      
\usepackage{lscape}         

\usepackage{booktabs}       
\usepackage{longtable}      
\usepackage{multirow}       
\usepackage{tabularx}       
\usepackage{tabulary}       
\usepackage{tabularray}     

\usepackage{algorithm}      
\usepackage{algpseudocode}  

\usepackage{xcolor}         
\usepackage{soul}           

\usepackage{url}            
\usepackage{hyperref}       
\hypersetup{
    colorlinks   = true,
    linkcolor    = blue,
    urlcolor     = blue,
    citecolor    = teal
}

\theoremstyle{thmstyleone}%
\theoremstyle{thmstyletwo}%

\theoremstyle{thmstylethree}%

\begin{document}

\title[Article Title]{Benchmarking Binary Classifiers Under Class Imbalance Without Rebalancing Techniques}


\author[1]{\fnm{Ali} \sur{Nawaz}}\email{700040004@uaeu.ac.ae}

\author*[1]{\fnm{Amir} \sur{Ahmad}}\email{amirahmad@uaeu.ac.ae}

\author[2]{\fnm{Shehroz} \sur{S. Khan}}\email{Shehroz.Khan@aum.edu.kw }

\affil[1]{\orgdiv{Department of Information Systems and Security, College of Information Technology and Center for Artificial Intelligence and Digital Innovation}, \orgname{United Arab Emirates University, United Arab Emirates}, \orgaddress{\city{Al Ain}, \country{UAE}}}

\affil[2]{\orgdiv{College of Engineering and Technology}, \orgname{American University of the Middle East}, \orgaddress{\city{Egaila}, \country{54200, Kuwait}}}


\abstract{Class imbalance poses a significant challenge to supervised classification, particularly in critical domains like medical diagnostics and anomaly detection where minority class instances are rare. While numerous studies have explored rebalancing techniques to address this issue, less attention has been given to evaluating the performance of binary classifiers under imbalance when no such techniques are applied. Therefore, the goal of this study is to assess the performance of binary classifiers "as-is", without performing any explicit rebalancing. Specifically, we systematically evaluate the robustness of a diverse set of binary classifiers across both real-world and synthetic datasets, under progressively reduced minority class sizes, using one-shot and few-shot scenarios as baselines. Our approach also explores varying data complexities through synthetic decision boundary generation to simulate real-world conditions. In addition to standard classifiers, we include experiments using undersampling, oversampling strategies, and one-class classification (OCC) methods to examine their behavior under severe imbalance. The results confirm that classification becomes more difficult as data complexity increases and the minority class size decreases. While traditional classifiers deteriorate under extreme imbalance, advanced models like TabPFN and boosting-based ensembles retain relatively higher performance and better generalization compared to traditional classifiers. Visual interpretability and evaluation metrics further validate these findings. Our work offers valuable guidance on model selection for imbalanced learning, providing insights into classifier robustness without dependence on explicit rebalancing techniques.
}




\maketitle

\section{Introduction}
Supervised classification is the process of mapping input data to predefined classes through mathematical modelling \cite{richards2022supervised}. Classification has many useful applications, including disease detection, spam detection, speech recognition, information retrieval, and numerous other fields \cite{shetty2022supervised}. Most supervised classification algorithms infer a function from labelled training data, consisting of training examples available either in raw form such as images or as meaningful extracted features.

Supervised methods generally perform better when data objects in different classes such as disease vs. healthy are evenly distributed \cite{li2022targeted}. However, if the distribution of data objects is skewed, classification decisions may be disproportionately influenced by the majority class, resulting in unreliable and biased outcomes \cite{feng2024fair}. This issue of class imbalance is critical, particularly in scenarios such as medical diagnostics \cite{feng2024fair}, fraud detection \cite{kennedy2024synthesizing}, and anomaly detection \cite{samariya2023comprehensive}, where accurate identification of minority class instances carries significant practical implications.

Identifying unusual behaviours or patterns in big datasets is an important research problem, as it may directly impact the health, safety, and economics of people and organizations \cite{gupta2018big}. While collecting data for normal behaviours or activities is relatively straightforward, obtaining data on abnormal, unusual, or anomalous behaviours can be challenging \cite{tay2023review} \cite{khan2014one}. Abnormal situations often occur rarely, infrequently, and diversely, such as failures of atomic power plants, rare diseases, or hacking incidents in financial markets. Consequently, representative samples of these abnormal events are typically scarce or entirely absent during the training phase of classification algorithms, but can still appear during testing. Supervised classifiers cannot handle such scenarios, as they are only trained on known classes. Many real-world datasets contain very few abnormal data points, known as imbalanced datasets \cite{haixiang2017learning}. In extreme cases, datasets may have no abnormal data points available for training \cite{khan2017detecting}, creating one-class datasets \cite{khan2014one}. Both imbalanced and one-class classification scenarios occur naturally across various application domains, including network security, finance, health, manufacturing, and Industry 4.0 \cite{hayashi2022one}. From a machine learning perspective, training traditional supervised classifiers to identify normal and abnormal situations is challenging due to the skewed data distribution.

To address the challenge of class imbalance, explicit rebalancing techniques such as oversampling, undersampling \cite{carvalho2025resampling}, and synthetic data generation methods (e.g., Synthetic Minority Over-sampling Technique (SMOTE)) are commonly used  \cite{chawla2002smote}. However, these explicit methods have their drawbacks. Synthetic data generation may distort the original data distribution, introducing misleading information or noise that can overfit model generalization \cite{ren2023systematic}. Undersampling methods risk losing critical information by removing potentially valuable majority class instances, leading to underfitting a model  \cite{yuan2023review}. Furthermore, these techniques often involve significant computational overhead, limiting their practicality in large-scale or real-time scenarios. The other possibility is to re-weight the binary class(es); however, that may only shift the operating threshold and may not learn a generalizable classification boundary.

Despite extensive research on explicit rebalancing methods, there remains a notable gap in understanding the inherent ability of binary classifiers to effectively handle imbalanced datasets without relying on such techniques. Few studies have systematically evaluated classifier performance under extreme imbalance conditions \cite{khan2016classification}, particularly in one-shot or few-shot learning scenarios where the minority class is severely underrepresented. Moreover, the intrinsic capacity of ensemble methods to manage such imbalances without explicit intervention has not been thoroughly explored \cite{ochal2023few}. Addressing this gap is crucial for gaining deeper insights into classifier resilience, simplifying methodological pipelines, and enhancing our understanding of decision boundaries in complex class distributions. Additionally, we also aim to determine the minimum possible imbalance ratio that enables reliable classification and generalization, while also providing clarity on what degree of class ratio qualifies as imbalanced and how it relates to imbalance and one-class classification strategies.

Therefore, in this study, we systematically investigate the performance of binary classifiers in highly imbalanced scenarios without applying explicit rebalancing techniques. By evaluating model performance across a range of imbalance ratios, including extreme cases with minimal minority class representation, and by employing ensemble learning approaches, synthetic benchmark datasets, and visual interpretability techniques \cite{ahmad2025one}, we aim to assess both the viability and limitations of classifiers under severe imbalance. Our approach also includes the generation of datasets that reflect increasing levels of non-linearity and structural complexity to simulate real-world classification challenges. By doing so, we aim to generate synthetic datasets representing diverse data complexities and geometric structures, then apply binary classifiers on the generated datasets under systematically reduced minority class sizes to determine the minimum possible imbalance ratio that enables reliable classification and generalization.

The main objectives of the study are;
\begin{itemize}
    \item To evaluate the robustness of traditional supervised classifiers, ensemble models, and deep tabular methods by systematically reducing minority class instances, including baseline cases with one-class classification (OCC), and analyzing their performance under increasing imbalance severity up to one-shot and few-shot scenarios.
    
    \item To identify the minimum imbalance ratio at which reliable classification and generalization are still achievable by using synthetic datasets that represent diverse data complexities and geometric structures, and by evaluating classifier robustness on these datasets through the application of binary classifiers under progressively reduced minority class sizes. 
            
\end{itemize}

The rest of this paper is organized as follows: Section 2 presents the related work. Section 3 describes the experimental methodology. Section 4 discusses the experimental results and key findings. Finally, Section 5 concludes the paper with final remarks.

\section{Related Work}
Imbalanced data classification has emerged as a critical challenge in machine learning, particularly in domains where the minority class represents rare but significant instances \cite{ghosh2024class} such as fraud detection, medical diagnosis, network intrusion, and fault monitoring. Traditional supervised learning algorithms often assume that data samples are equally distributed across classes; however, in real-world scenarios, this assumption rarely holds. The dominance of the majority class samples can bias the learning process, resulting in poor detection of minority instances \cite{sarker2021machine}. This issue becomes even more prominent in extreme cases such as one-shot or few-shot learning in which only a handful or even a single minority instance is available for training \cite{tyukin2021demystification}. Consequently, a wide range of techniques have been proposed to address the class imbalance problem, spanning data-level resampling \cite{carvalho2025resampling}, algorithm-level modifications \cite{fernandez2018algorithm}, and hybrid frameworks \cite{galar2011review}. This section reviews recent and foundational work in imbalanced data classification also summarized in Table \ref{review}.

Thabtah et al. (2020) \cite{thabtah2020data} investigated the impact of imbalance ratios on classification accuracy using a Naive Bayes classifier. Their empirical evaluation across a range of imbalance levels and real datasets indicated that increased imbalance degrades minority class accuracy, even if overall accuracy remains high. They explored the effectiveness of oversampling and cost-sensitive adjustments but noted the need for further exploration into intrinsic classifier performance.

 Khan et al. (2024) \cite{khan2024review} present a comprehensive review of methodologies addressing class imbalance, organizing the discussion around preprocessing (data-level), algorithm-level, and hybrid strategies. They evaluate undersampling such as Tomek links, oversampling such as SMOTE, ADASYN, and synthetic data generation \cite{chawla2002smote}, highlighting their strengths and weaknesses. Algorithm-level approaches include cost-sensitive learning, threshold-moving, and boosting variations like RUSBoost. They also highlight the effectiveness of hybrid methods like SMOTEBoost \cite{chawla2003smoteboost}. Although the paper does not present new experimental data, it surveys applications from UCI and KDD datasets \cite{UCI}. It calls for improvements in method generalizability, scalability, and application to real-world domains involving image, text, and time series data

Gao et al. (2025) \cite{gao2025comprehensive} provided a comprehensive survey on techniques to handle imbalance in different data formats. Their taxonomy spans data-level (e.g., SMOTE, GANs), algorithm-level (e.g., re-weighted loss), and hybrid methods. They emphasized the growing importance of deep learning methods and pointed out the necessity to develop targeted imbalance strategies for complex data such as images, text, and graphs. Aguiar et al. (2024) \cite{aguiar2024survey} conducted an exhaustive survey and empirical study on imbalanced data streams. They presented a taxonomy of stream learning algorithms, proposed a benchmark of 515 scenarios with static and dynamic imbalance, and evaluated 24 algorithms. Their findings underscore the difficulty of handling both concept drift and imbalance simultaneously and advocate for ensemble-based and adaptive mechanisms tailored for evolving data distributions.

He and Garcia (2009) \cite{he2009learning} offered foundational insight into imbalanced learning techniques. They categorized the solutions into sampling, cost-sensitive learning, kernel-based approaches, and active learning. Their work provided theoretical and experimental justification for resampling methods but also highlighted limitations, such as overfitting in oversampling and information loss in undersampling. They advocated more research in high-dimensional and small-sample scenarios.

Polak et al. (2024) \cite{billion2024low} extended the discussion into the realm of few-shot learning (FSL), showing how class imbalance and low-data regimes intersect. They compared prototypical networks, meta-learning strategies, and generative models, concluding that many FSL methods inherently exhibit resilience to class imbalance. However, the literature lacks sufficient experiments validating these methods under extreme imbalance conditions.

Ochal et al. (2021) \cite{ochal2023few} focused on few-shot learning scenarios under class imbalance, analyzing task-level and dataset-level imbalance. Using benchmarks like Omniglot and MiniImageNet, they introduced random-shot meta-training and evaluated rebalancing losses. Their experiments demonstrated that common FSL models suffer under imbalance unless modified with oversampling or loss-weighting techniques.

Wang et al. \cite{wang2019dynamic} proposed Dynamic Curriculum Learning (DCL), a method combining curriculum learning with adaptive sampling and dynamic loss weighting to handle imbalance. Their framework consists of a sampling scheduler that evolves from imbalanced to balanced distribution and a loss scheduler that shifts emphasis between classification loss and metric learning loss. Evaluated on CelebA and RAP datasets, DCL outperformed baseline methods by balancing generalization and discrimination throughout training. Kamalov and Denisov (2020) \cite{kamalov2020gamma} proposed a gamma-distribution-based sampling method to improve oversampling effectiveness. Unlike SMOTE, which interpolates linearly, this method uses the asymmetric gamma distribution to create synthetic minority points close to real instances. Evaluated across 24 datasets, their method consistently outperformed SMOTE and ADASYN in F1-score and AUC, especially under high imbalance.

While these studies have focused extensively on overcoming imbalance through various forms of resampling, loss reweighting, or architectural modification, our work diverges by eliminating any external intervention and assessing classifiers “as-is.” We provide a rigorous evaluation of standard classifiers under decreasing minority class presence including one-shot and few-shot scenarios on both real-world and synthetic datasets. Moreover, we apply ensemble models such as BaggingRF \cite{altman2017ensemble} and RUSBoost \cite{seiffert2009rusboost}, not to rebalance data but to assess their natural resilience to imbalance. Through decision boundary visualization, we add interpretability to our findings, offering unique insight into how classifiers separate classes under extreme imbalance.

\section{Our Proposed Approach}

Our main objective is to assess the intrinsic robustness of binary classifiers when handling class imbalance without explicit rebalancing techniques. To systematically achieve this goal, we employ three key experimental strategies:

\subsection{Systematic Reduction}

We reduced the minority class datapoints highlighted in Step~1 to observe its effects on the performance of various ML classifiers. To simulate increasing levels of class imbalance, we used six different proportions of the abnormal class: 100\%, 50\%, 25\%, 10\%, 5\%, and 1\%. This means that in each cross-validation fold, we retained all the minority class samples in the 100\% setting, half in the 50\% setting, a quarter in the 25\% setting, and so on, reducing further down to only 1\% of the original minority samples. For percentage values below 1, we retain a single representative minority datapoint for training. For each level, the majority class datapoints were kept intact, while only the minority samples were reduced in size. This helped us systematically analyze the sensitivity of each model to extreme imbalance scenarios. 

\begin{landscape}

\begin{longtable}{|>{\hspace{0pt}}m{0.1\linewidth}|
>{\centering\hspace{0pt}}m{0.09\linewidth}|
>{\centering\hspace{0pt}}m{0.07\linewidth}|
>{\centering\hspace{0pt}}m{0.07\linewidth}|
>{\centering\hspace{0pt}}m{0.07\linewidth}|
>{\centering\hspace{0pt}}m{0.07\linewidth}|
>{\centering\hspace{0pt}}m{0.08\linewidth}|
>{\hspace{0pt}}m{0.15\linewidth}|}
\caption{Comparison of our work with related work}
\label{review} \\
\hline
\textbf{Paper / Our Work} & \textbf{Systematic Reduction} & \textbf{One-Shot / Few-Shot} & \textbf{Ensemble Methods} & \textbf{Synthetic Data Generation} & \textbf{One-Class Classification Models} & \textbf{Decision Boundary Visualization} & \textbf{Evaluation Metrics} \\
\hline
\endfirsthead
\hline
\textbf{Paper / Our Work} & \textbf{Systematic Reduction} & \textbf{One-Shot / Few-Shot} & \textbf{Ensemble Methods} & \textbf{Synthetic Data Generation} & \textbf{One-Class Classification Models} & \textbf{Decision Boundary Visualization} & \textbf{Evaluation Metrics} \\
\hline
\endhead
Thabtah \emph{et al.} (2019) \cite{thabtah2020data}  & \checkmark & & & & & & AUC-ROC , Precision, Recall, F1 \\
\hline
Khan \emph{et al.} (2024) \cite{khan2024review}  & & & \checkmark & \checkmark & & & AUC-ROC , Accuracy \\
\hline
Gao \emph{et al.} (2025) \cite{gao2025comprehensive} &  & & \checkmark & \checkmark & & & AUC-ROC , F1 \\
\hline
Aguiar et al. (2024) \cite{aguiar2024survey} & \checkmark & & \checkmark & & & & AUC-ROC , G-mean \\
\hline
He \& Garcia (2009) \cite{he2009learning} & \checkmark & & \checkmark & \checkmark & & & AUC-ROC , F1 \\
\hline
Polak et al. (2022) \cite{billion2024low} & & \checkmark & & & & & AUC-ROC , Accuracy \\
\hline
Ochal et al. (2023) \cite{ochal2023few} & & \checkmark & & & & \checkmark & AUC-ROC , G-mean \\
\hline
Wang et al. (2019) \cite{wang2019dynamic} & & & & \checkmark & & \checkmark & AUC-ROC , G-mean \\
\hline
Kamalov \& Denisov (2020) \cite{kamalov2020gamma} & & & & \checkmark &  & \checkmark & AUC-ROC , Boundary Plots \\
\hline
\textbf{Our Work} & \checkmark & \checkmark & \checkmark & \checkmark & \checkmark & \checkmark & AUC-ROC , AUC-PR, Accuracy, Precision, Recall, F1-score, G-mean \\
\hline
\end{longtable}

\end{landscape}

\vspace{0.4 ex}
\noindent\rule{\linewidth}{0.5pt} 

\noindent\textbf{Step 1} Systematic Abnormal Class Reduction for Model Evaluation

\noindent\rule{\linewidth}{0.5pt} 
\vspace{0.4ex}

\begin{algorithmic}[1]
\Require Labeled dataset $D = \{(\mathbf{x}_i, y_i)\}_{i=1}^N$ with $y_i \in \{0,1\}$
\Require Percentages $P = \{100, 50, 25, 10, 5, 1\}$
\Require Classifier set $\mathcal{C}$
\Require Number of cross-validation folds $K$
\Ensure Performance metrics for each classifier, each percentage, and each fold

\For{fold $k = 1$ to $K$}
    \State Split $D$ into training set $D_{\text{train}}^{(k)}$ and test set $D_{\text{test}}^{(k)}$
    \State $D_{\text{train, normal}}^{(k)} \gets \{(\mathbf{x}_i, y_i) \in D_{\text{train}}^{(k)} : y_i = 0\}$
    \State $D_{\text{train, abnormal}}^{(k)} \gets \{(\mathbf{x}_i, y_i) \in D_{\text{train}}^{(k)} : y_i = 1\}$
    \For{each percentage $p$ in $P$}
        \If{$p = 100$}
            \State $D_{\text{train,red}}^{(k,p)} \gets D_{\text{train}}^{(k)}$
        \Else
            \State Randomly select $S_{\text{abn}}^{(k,p)} \subseteq D_{\text{train, abnormal}}^{(k)}$ of size $\max\{1, \lfloor |D_{\text{train, abnormal}}^{(k)}| \cdot p / 100 \rfloor\}$
            \State $D_{\text{train,red}}^{(k,p)} \gets D_{\text{train, normal}}^{(k)} \cup S_{\text{abn}}^{(k,p)}$
        \EndIf
        \State Shuffle $D_{\text{train,red}}^{(k,p)}$
        \For{each classifier $c$ in $\mathcal{C}$}
            \State Train $c$ on $D_{\text{train,red}}^{(k,p)}$
            \State Evaluate $c$ on $D_{\text{test}}^{(k)}$ and record performance metrics
        \EndFor
    \EndFor
\EndFor
\State Mean and standard deviation of each performance metric
\end{algorithmic}

\subsection{One-shot and Few-shot}

To investigate the performance of binary classifiers in extremely imbalanced settings, we employed a systematic strategy for constructing one-shot and few-shot training sets, as summarized in Step~2 \cite{ochal2023few}. We consider this setup as our baseline for evaluating classifier robustness. In this procedure, we explore three scenarios with $k = 1$, $3$, and $5$ minority class datapoints retained in each fold of cross-validation. For every value of $k$, we create reduced training sets by retaining all majority class examples and randomly sampling only $k$ minority class instances. These minimal datasets simulate real-world situations where obtaining minority samples is difficult. To ensure the robustness and generalizability of the results in one-shot and few-shot settings, we repeated the experiments ten times using different random selections of minority samples and reported the average of AUC-ROC and AUC-PR scores. This repetition mitigates the risk of results being biased or overly specific to a particular choice of a minority samples. Each classifier is trained on the reduced set and tested on the full test fold to evaluate performance under extreme imbalance.

\vspace{0.4 ex}
\noindent\rule{\linewidth}{0.5pt} 

\noindent\textbf{Step 2} One-Shot and Few-Shot ($k=1,3,5$) Training Set Construction

\noindent\rule{\linewidth}{0.5pt} 
\vspace{0.4ex}

\begin{algorithmic}[1]
\Require Labeled dataset $D = \{(\mathbf{x}_i, y_i)\}_{i=1}^N$, $y_i \in \{0,1\}$
\Require Set of desired minority samples $K = \{1,3,5\}$
\Require Classifier set $\mathcal{C}$
\Require Number of cross-validation folds $F$

\For{$k$ in $K$}
    \For{each cross-validation fold $f=1$ to $F$}
        \State Partition $D$ into training set $D_{\text{train}}^{(f)}$ and test set $D_{\text{test}}^{(f)}$
        \State Identify all normal indices: $I_{0} = \{i : y_i = 0\}$
        \State Identify all abnormal indices: $I_{1} = \{i : y_i = 1\}$
        \If{$|I_1| < k$}
            \State \textbf{continue} to next fold
        \EndIf
        \State Randomly sample $k$ abnormal indices: $S_{1} \subset I_{1}$, $|S_{1}| = k$
        \State Construct reduced training set: $D_{\text{train}}^{(f,k)} = \{(\mathbf{x}_i, y_i): i \in I_0\} \cup \{(\mathbf{x}_i, y_i): i \in S_{1}\}$
        \State Shuffle $D_{\text{train}}^{(f,k)}$
        \For{each classifier $c$ in $\mathcal{C}$}
            \State Train $c$ on $D_{\text{train}}^{(f,k)}$
            \State Evaluate $c$ on $D_{\text{test}}^{(f)}$ and record performance metrics
        \EndFor
    \EndFor
    \State Mean and standard deviation of each performance metrics
\EndFor
\end{algorithmic}

\noindent\rule{\linewidth}{0.4pt} 

\subsection{Synthetic Decision Boundary Generation}

\begin{figure*}[htbp]
    \centering

    \subfigure[Linear decision boundary synthetic dataset]{\includegraphics[width=0.45\textwidth]{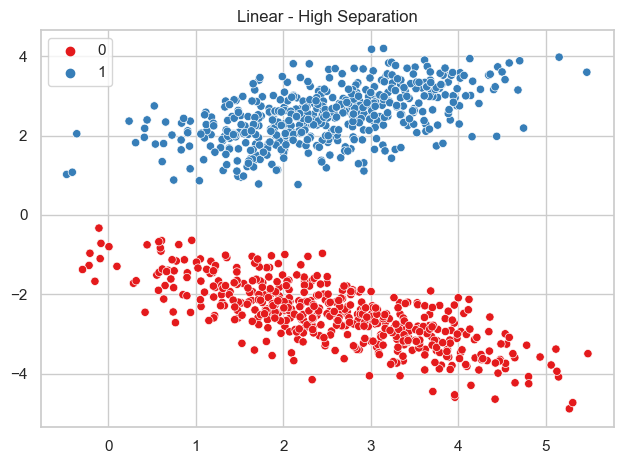}} \hfill
    \subfigure[Moderate Non-Linear decision boundary synthetic dataset]{\includegraphics[width=0.45\textwidth]{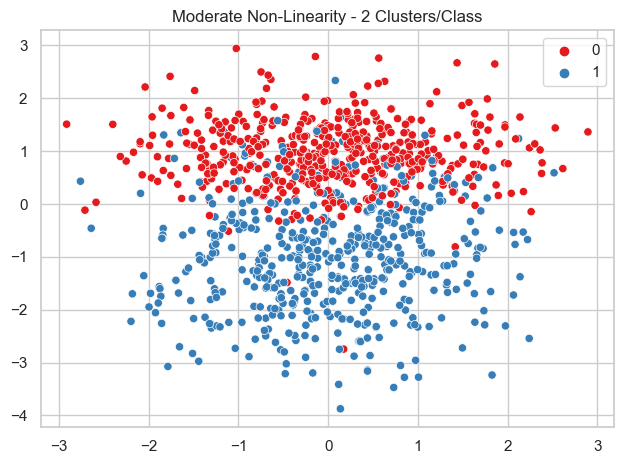}} \\[1ex]

    \subfigure[Non-Linear-Redundancy decision boundary synthetic dataset]{\includegraphics[width=0.455\textwidth]{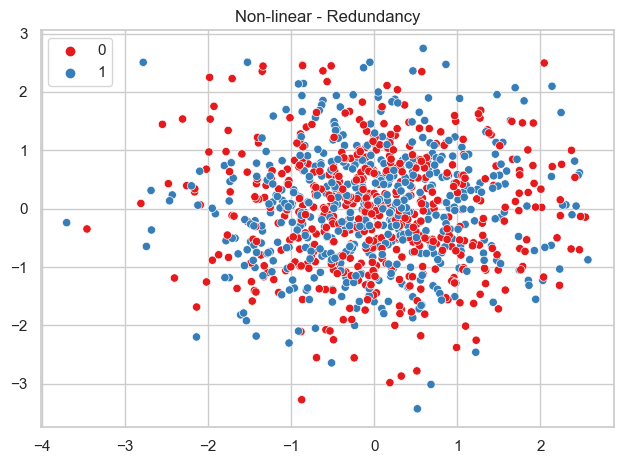}} \hfill
    
    \subfigure[Non-Linear-Gaussian quantiles decision boundary synthetic dataset]
    {\includegraphics[width= 0.455\textwidth]{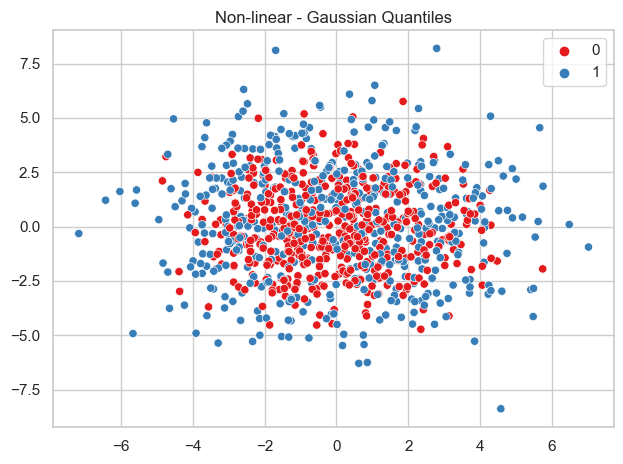}} 
    \subfigure[Hard Non-linear - XOR decision boundary synthetic dataset]
    {\includegraphics[width= 0.455\textwidth]{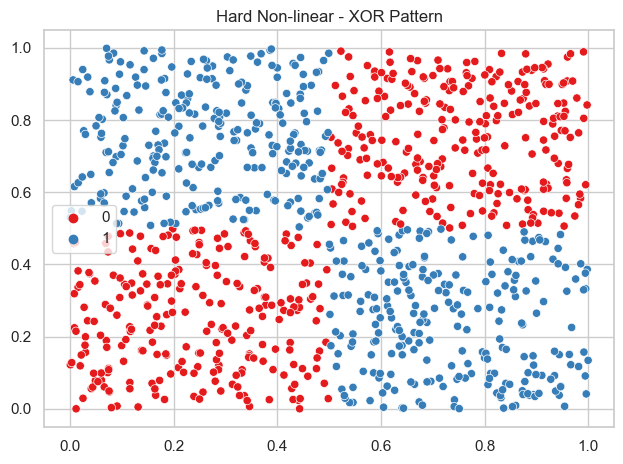}} \hfill
    \\
     \caption{Generated synthetic decision boundary datasets}
     \label{synthetic boundary}
\end{figure*}

To investigate the response of binary classifiers under varying data complexities and imbalance scenarios, we generated a suite of synthetic benchmark datasets, as shown in Figure \ref{synthetic boundary}. These datasets were designed to represent diverse classification challenges, including linearly separable distributions subfigure Figure \ref{synthetic boundary} (a), moderately non-linear clusters with noise Figure \ref{synthetic boundary} (b), high-dimensional redundancy with PCA-based dimensionality reduction Figure \ref{synthetic boundary} (c), circular class structures via Gaussian quantiles Figure \ref{synthetic boundary} (d), and a hard non-linear XOR pattern Figure \ref{synthetic boundary} (e) \cite{thanathamathee2013handling}. Each dataset simulates a distinct decision boundary complexity to evaluate classifier behavior across a range of synthetic scenarios.

Following the creation of these datasets, we systematically reduced the proportion of minority class examples in the training sets to simulate real-world class imbalance. We considered six reduction levels: 100\%, 50\%, 25\%, 10\%, 5\%, and 1\% of the original minority class samples. At each reduction level, we evaluated the ability of various binary classifiers to learn and generalize, with the primary objective of determining the minimum possible imbalance ratio that enables reliable classification and generalization. This analysis not only sheds light on classifier performance across imbalance scenarios but also helps reveal how geometric complexity in the data interacts with skewed class distributions, providing deeper insight into model behavior under imbalanced conditions.

\section{Experimental Setup}
In this section, we provide detailed descriptions of the datasets utilized, the machine learning models applied in our experiments, and the evaluation procedures employed to assess classifier performance comprehensively.

\subsection{Datasets}

The datasets used in the experimentation are listed and summarized in Table \ref{datasets}. The public datasets are acquired from a UCI \cite{UCI} and Kaggle repository \cite{Kaggle}.

\begin{table*}[t]   
\centering
\caption{Summary of datasets and their respective imbalance ratios}
\label{datasets}
\begin{tabular}{|c|l|c|r|r|r|}
\hline
\textbf{S. No.} & \textbf{Dataset} & \textbf{Short Form} & \textbf{Majority Points} & \textbf{Minority Points} & \textbf{Imbalance Ratio} \\
\hline
1  & Breast Cancer     & D1  & 357  & 8    & 0.0224 \\
2  & Pen Local         & D2  & 6712 & 10   & 0.0015 \\
3  & Pen Global        & D3  & 718  & 89   & 0.1240 \\
4  & Letter            & D4  & 1498 & 100  & 0.0667 \\
5  & Annthyroid        & D5  & 6666 & 248  & 0.0372 \\
6  & Satellite         & D6  & 5025 & 73   & 0.0145 \\
7  & Glass             & D7  & 163  & 51   & 0.3135 \\
8  & Segment           & D8  & 1979 & 329  & 0.1662 \\
9  & Pima              & D9  & 500  & 268  & 0.5360 \\
10 & Yeast4            & D10 & 1433 & 51   & 0.0356 \\
11 & Yeast5            & D11 & 1440 & 44   & 0.0306 \\
12 & Yeast6            & D12 & 1449 & 35   & 0.0241 \\
13 & Abalone           & D13 & 4142 & 32   & 0.0077 \\
14 & Abalone9-18       & D14 & 689  & 42   & 0.0609 \\
15 & Ecoli4            & D15 & 316  & 20   & 0.0633 \\
16 & PC1               & D16 & 1032 & 77   & 0.0746 \\
17 & CM1               & D17 & 449  & 49   & 0.1091 \\
18 & KC1               & D18 & 1783 & 326  & 0.1829 \\
19 & KC2               & D19 & 415  & 107  & 0.2578 \\
\hline
\end{tabular}
\end{table*}

\subsection{Machine Learning Models}

We applied a diverse set of binary classifiers listed in the Table \ref{tab:classifier_details}, including traditional models such as DT \cite{quinlan1996learning}, k-NN \cite{xu2022outlier}, ensemble techniques (such as RF, XGBoost, LGBM, CatBoost, BaggingRF, and RUSBoost) \cite{galar2011review}, margin-based classifiers (SVM) \cite{cervantes2020comprehensive}, and an advanced transformer-based TabPFN model \cite{hollmann2025accurate}.

\begin{table*}[h!]
\centering
\caption{Summary of classifiers: definitions and applied key parameters}
\begin{tabular}{p{3.5cm} p{5.5cm} p{6cm}}
\toprule
\textbf{Classifiers} & \textbf{Definition} & \textbf{Set Parameters / Characteristics} \\
\midrule
\textbf{TabPFN} \cite{hollmann2025accurate} & A transformer-based probabilistic model designed for tabular data that approximates Bayesian posterior inference. & Supports single forward-pass inference; effective in low-data regimes and robust to noise. \\
\textbf{DecisionTree (DT)} \cite{james2023tree} & A classical tree-based model that recursively splits features to maximize information gain. & \texttt{criterion="gini"}, \texttt{max\_depth=None} \\
\textbf{RandomForest (RF)} \cite{breiman2001random} & An ensemble method that constructs multiple decision trees using bootstrapped samples and random feature selection. & \texttt{n\_estimators=100}, \texttt{max\_depth=None}, \texttt{bootstrap=True} \\
\textbf{Extreme Gradient Boosting (XGB)} \cite{chen2015xgboost} & An efficient gradient boosting framework that builds trees sequentially to reduce residual errors. & \texttt{n\_estimators=100}, \texttt{learning\_rate=0.1}, \texttt{max\_depth=6} \\
\textbf{Light Gradient Boosting Machine (LGBM)} \cite{fan2019light} & A histogram-based gradient boosting algorithm that performs leaf-wise tree growth for high efficiency and accuracy. & \texttt{boosting\_type="gbdt"}, \texttt{learning\_rate=0.1}, \texttt{num\_leaves=31} \\
\textbf{Support Vector Machine (SVM)} \cite{cervantes2020comprehensive} & A margin-based classifier that constructs the optimal hyperplane using kernel functions. & \texttt{kernel="rbf"}, \texttt{C=1.0}, \texttt{probability=True} \\
\textbf{k-Nearest Neighbors (k-NN)} \cite{cunningham2021k} & A non-parametric algorithm that classifies samples based on the majority vote of their k-nearest neighbors. & \texttt{n\_neighbors=5}, \texttt{metric="minkowski"} \\
\textbf{CatBoost} \cite{prokhorenkova2018catboost} & A gradient boosting algorithm with native handling of categorical features and ordered boosting. & \texttt{iterations=100}, \texttt{depth=6}, \texttt{learning\_rate=0.1}, \texttt{verbose=0} \\
\textbf{Bagging Random Forest (Bagging RF)} \cite{altman2017ensemble} & A bagging ensemble method using multiple Random Forests with ROS to improve model stability and accuracy. & \texttt{n\_estimators=5}, base estimator: \texttt{RFClassifier(n\_estimators=10)} \\
\textbf{Random Undersampling Boosting (RUSBoost)} \cite{seiffert2009rusboost} & A hybrid ensemble combining random undersampling with boosting, effective for imbalanced data. & \texttt{n\_estimators=5}, base estimator: \texttt{DTClassifier}, \texttt{sampling\_strategy="auto"} \\
\textbf{One-class support vector machine (OCSVM)} \cite{khan2014one} & A one-class SVM used for anomaly detection, learns the boundary of normal data in high-dimensional space. & \texttt{kernel="rbf"}, \texttt{nu=0.5} \\
\textbf{Isolation Forest} \cite{michael2023outlier} & An unsupervised anomaly detection method that isolates outliers by recursive partitioning. & \texttt{n\_estimators=100}, \texttt{contamination=0.1}, \texttt{random\_state=42} \\
\textbf{Local Outlier Factor (LOF)} \cite{xu2022outlier} & A density-based outlier detection algorithm measuring local deviation of a point with respect to its neighbors. & \texttt{n\_neighbors=20}, \texttt{contamination=0.1} \\
\bottomrule
\end{tabular}
\label{tab:classifier_details}
\end{table*}

\subsection{Performance Measures}

Experiments were conducted on a high-performance computing workstation, specifically the Titan GT77HX 13VI, equipped with an NVIDIA GeForce RTX 4090 GPU featuring 16 GB of dedicated memory. This powerful hardware configuration enabled efficient training and evaluation of the models across multiple datasets and experimental settings.

All classifiers were evaluated using a 2×5-fold stratified cross-validation strategy i.e., 5-fold cross-validation repeated twice to ensure effective and reliable performance estimates. This approach helps minimize the variance associated with individual splits and yields more stable average performance metrics. To evaluate the performance of classifiers, particularly in the context of imbalanced datasets, we employed a set of standard evaluation metrics. The final evaluation was based on the mean and standard deviation values of  AUC-ROC, AUC-PR, accuracy, precision, recall, F1-score and G-mean across all 10 folds \cite{salmanpour2025machine}.

\section{Results and Discussion}

In this section, we present detailed results and interpret classifier performance under various experimental scenarios. Complete results are provided in the Supplementary File.  
\subsection{Robustness under Systematic Reduction}
\begin{figure*}[htbp]
    \centering

    \subfigure[Breast Cancer  Datasets(D1)]{\includegraphics[width=0.45\textwidth]{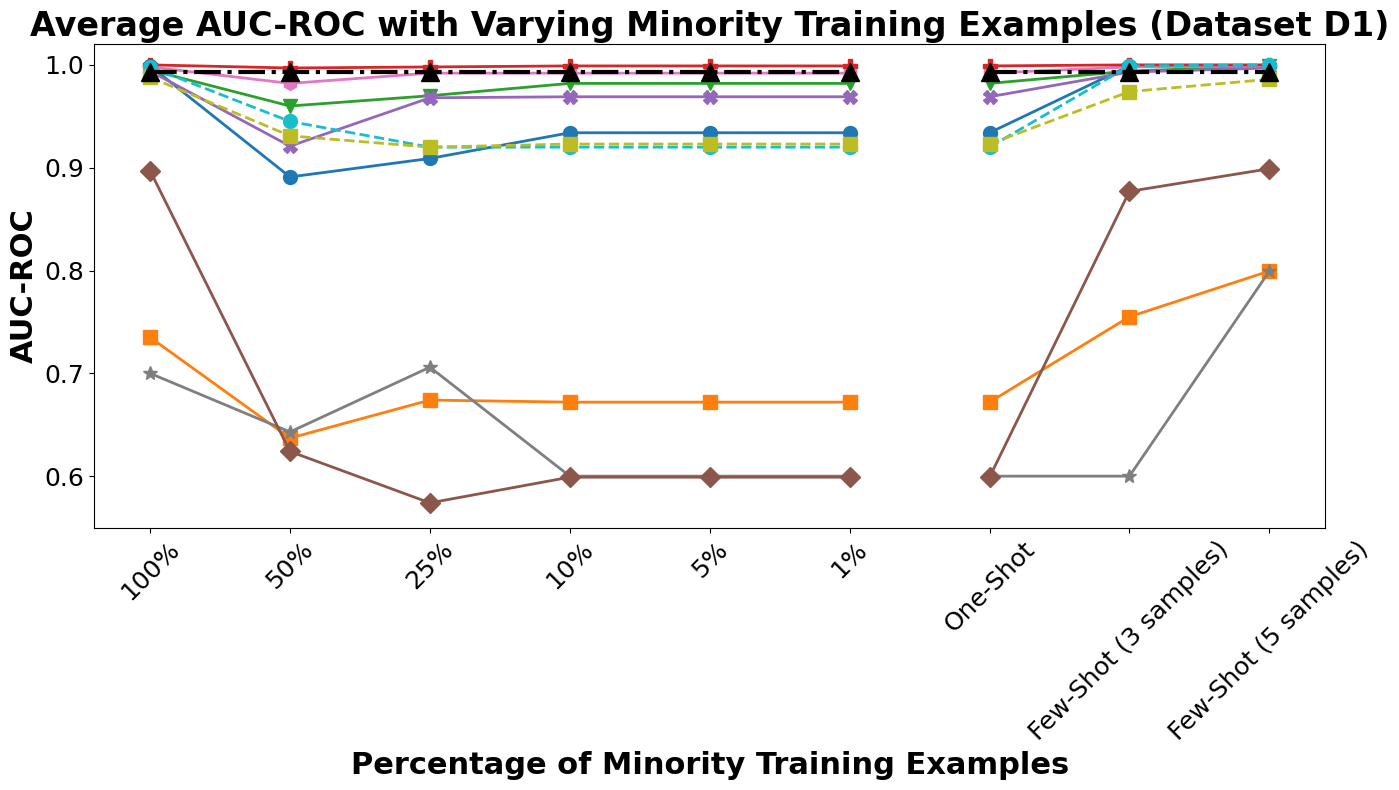}} \hfill
    \subfigure[Pen Local Datasets (D2)]{\includegraphics[width=0.45\textwidth]{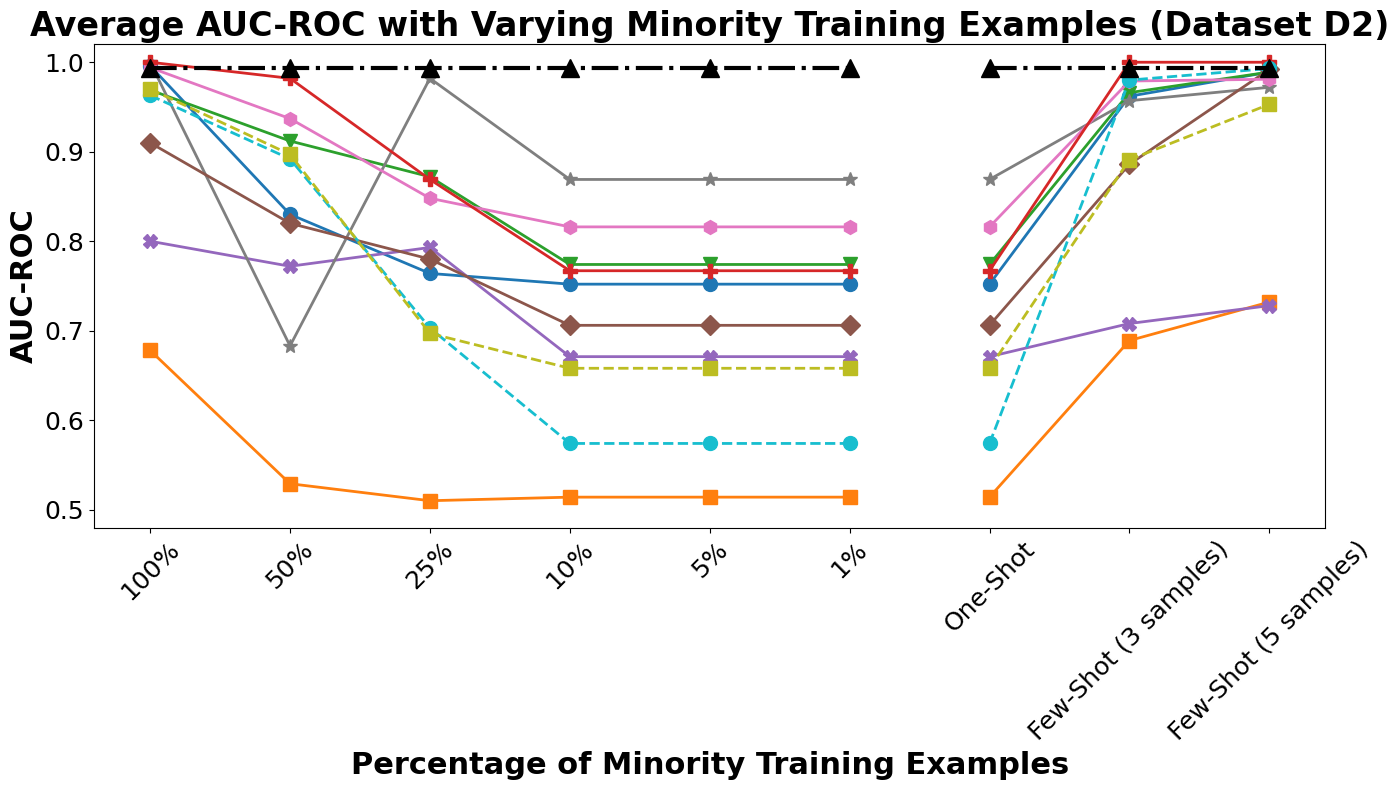}} \\[1ex]

    \subfigure[Pen Global Datasets (D3) ]{\includegraphics[width=0.45\textwidth]{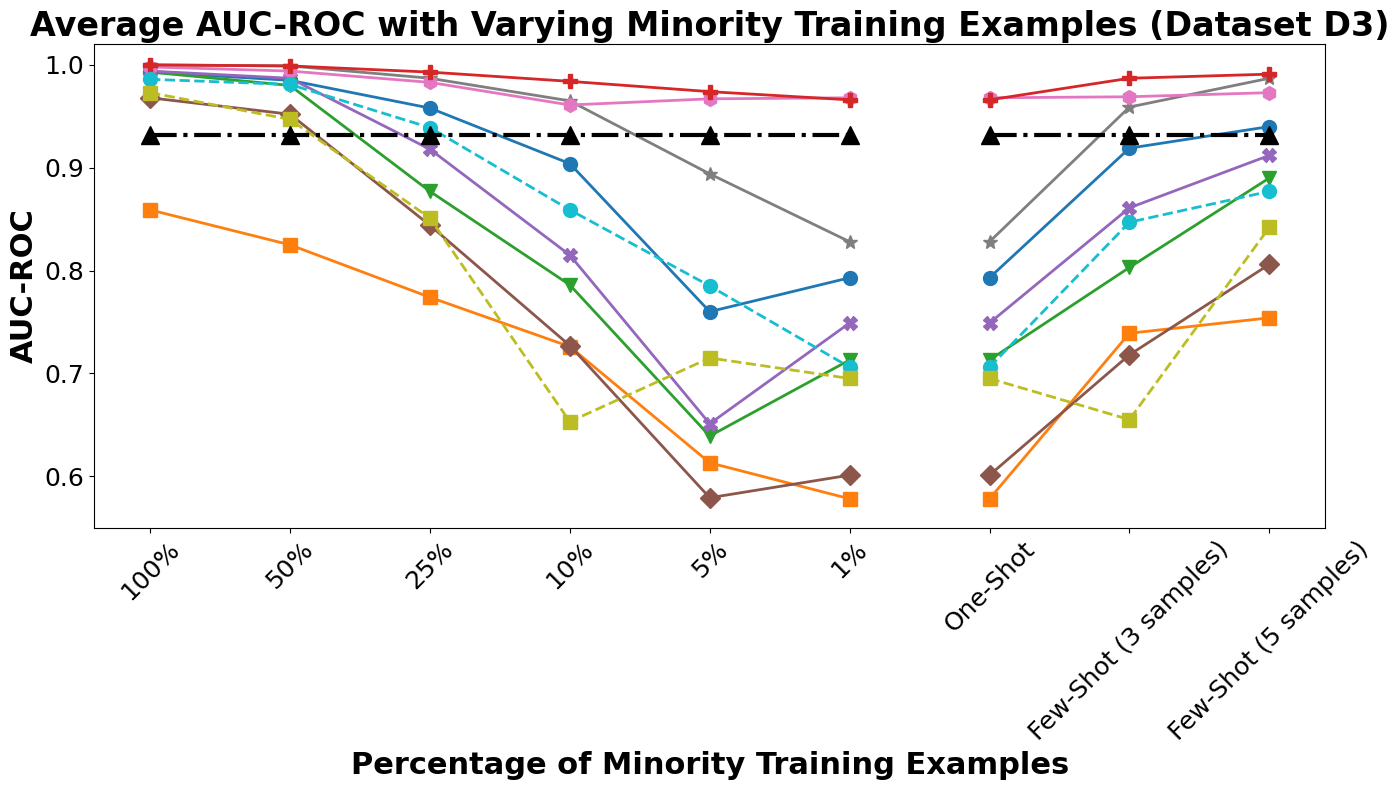}} \hfill
    \subfigure[Letter (D4)]{\includegraphics[width=0.45\textwidth]{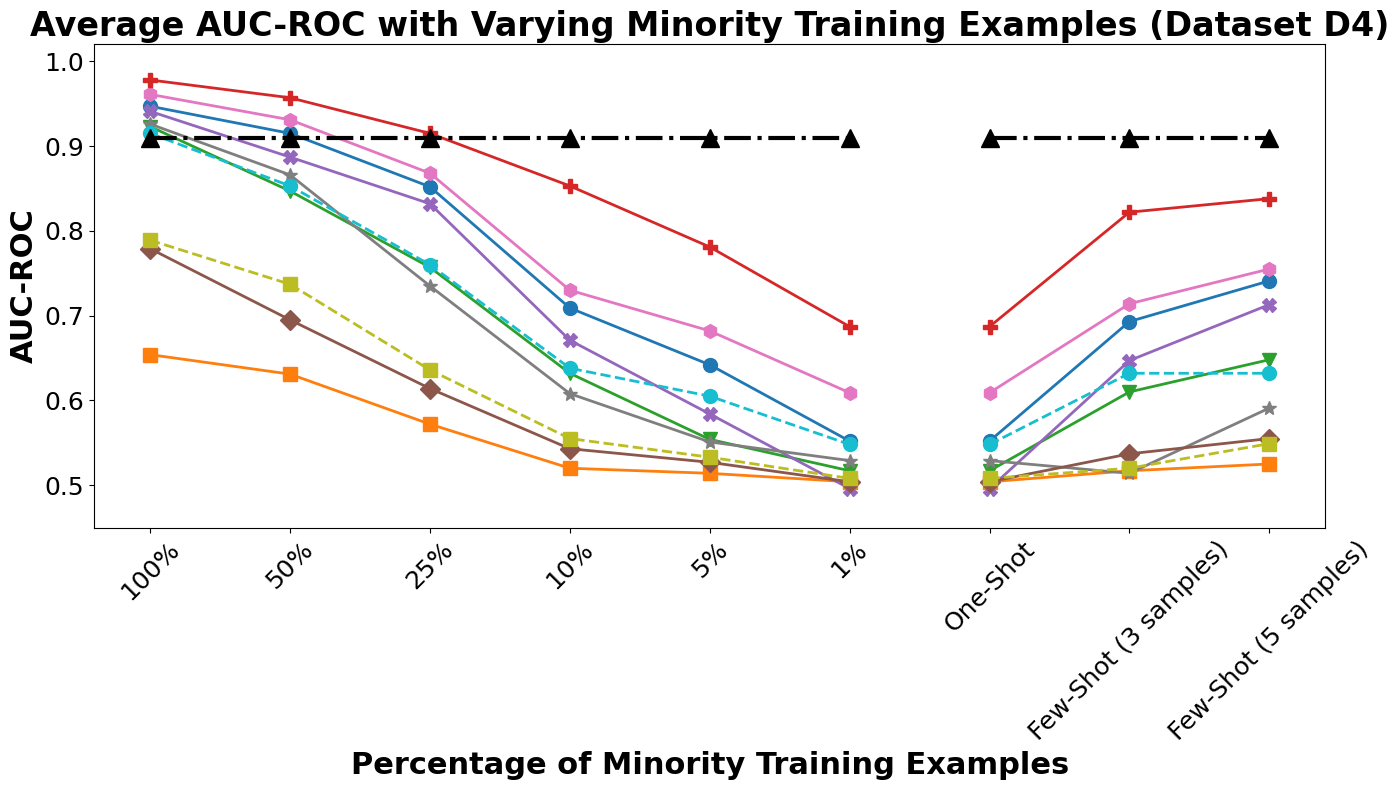}} \\[1ex]

    \subfigure[Annthyroid Datasets (D5)]{\includegraphics[width=0.45\textwidth]{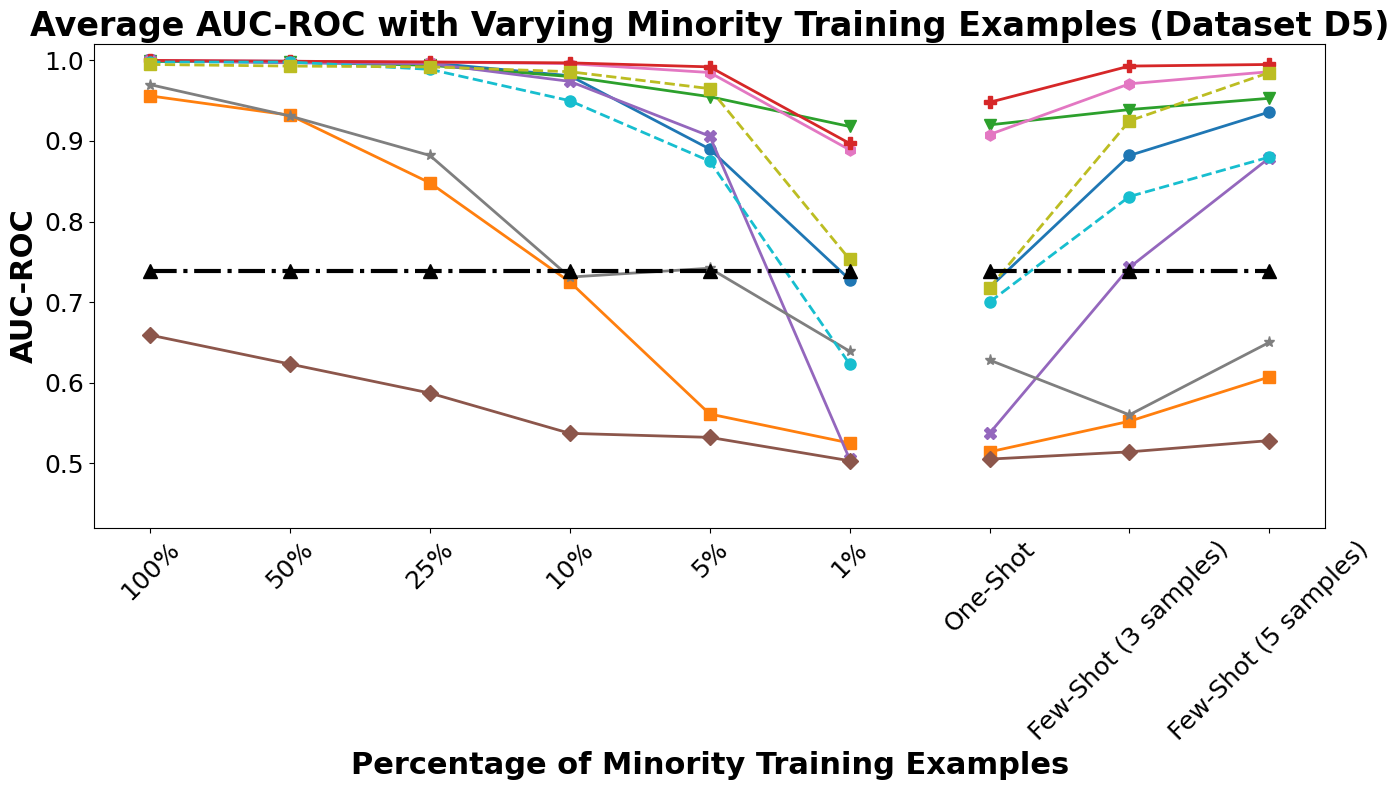}} \\[1ex]

    \caption{Average AUC-ROC with varying minority training datapoints of different datasets.}
    \label{auc-roc across different datasets}
\end{figure*}

\begin{figure}[htbp]\centering
     \includegraphics[width=\linewidth]{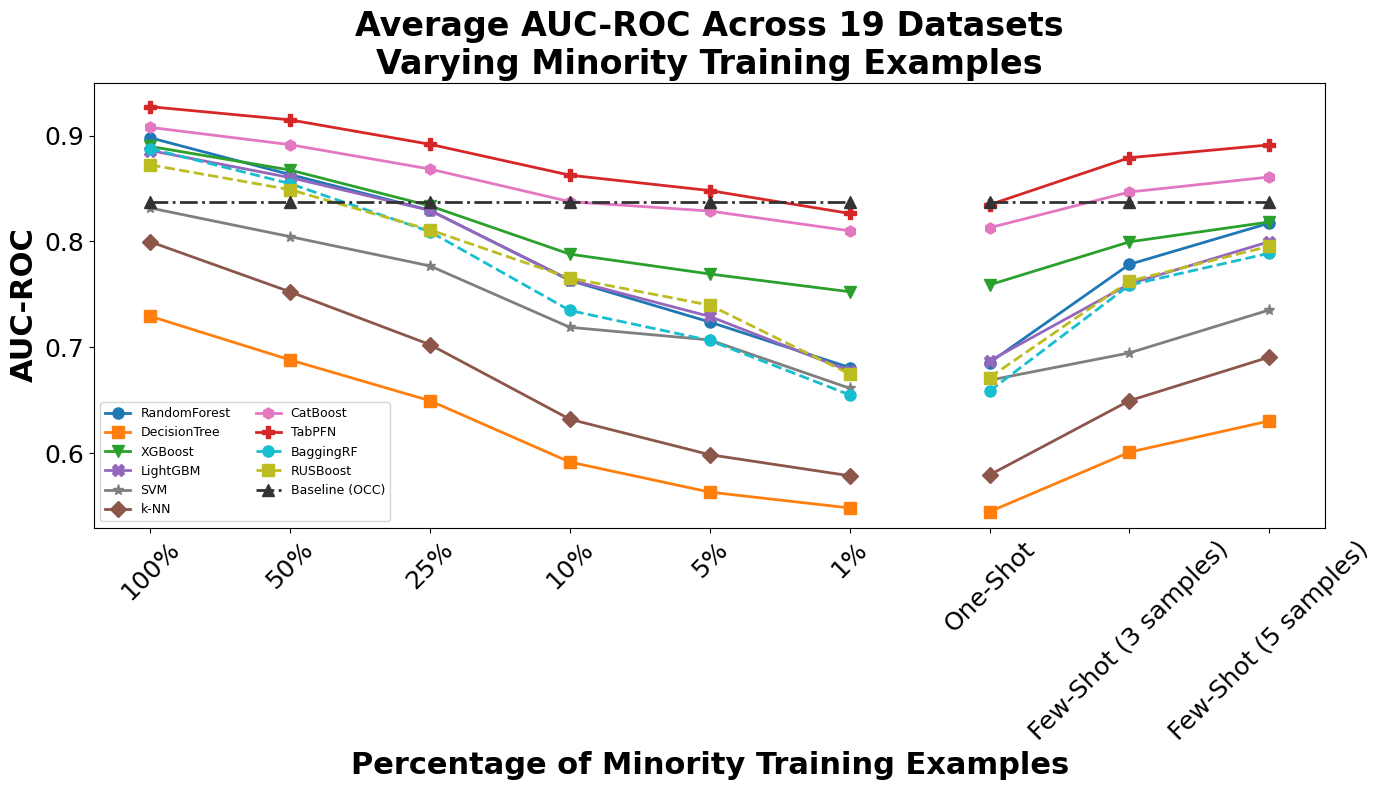}
  \caption{Average AUC-ROC across 19 datasets on minority training examples. X-axis represents percentage of minority training while Y-axis represents AUC-ROC}
 \label{Average AUC-ROC}
\end{figure}

\begin{figure}[htbp]\centering
     \includegraphics[width=\linewidth]{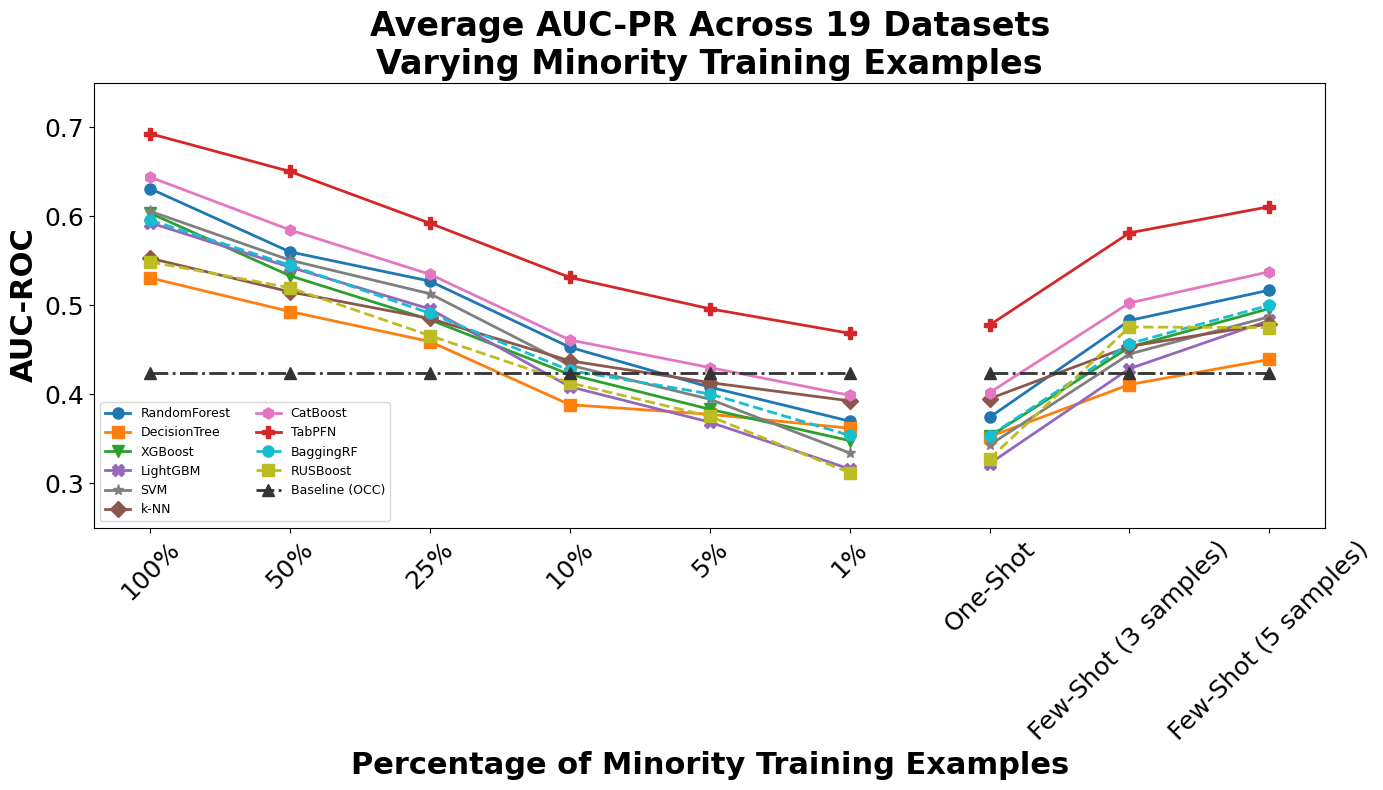}
  \caption{Average AUC-PR across 19 datasets on minority training examples. X-axis represents the percentage of minority training, while Y-axis represents AUC-PR.}
 \label{Average AUC-PR}
\end{figure}

Figures~\ref{Average AUC-ROC} and~\ref{Average AUC-PR} present the average AUC-ROC and AUC-PR scores, respectively, as the number of minority class training examples is systematically reduced. For space utilization, results are presented for five datasets in Figures~\ref{auc-roc across different datasets} and~\ref{auc-pr across different datasets}, while the complete results are provided in the Supplementary file. 

As the proportion of minority class examples decreases, most classifiers reveal a consistent drop in both AUC-ROC and AUC-PR scores. This decline becomes particularly severe below the 25\% and worsens further at 10\%, 5\%, and especially 1\%. These trends confirm that standard classifiers such as Decision Tree (DT), k-Nearest Neighbors (k-NN), Random Forest (RF), and XGBoost struggle to maintain performance under high imbalance without any explicit balancing strategies.

Notably, certain models demonstrate strong stability in these constrained settings. TabPFN consistently delivers the highest performance across all levels of minority reduction, including extreme imbalance, confirming its robustness in data-scarce scenarios. Similarly, CatBoost and LightGBM (LGBM) maintain relatively stable performance down to 10\% minority representation, further illustrating their adaptability under skewed class distributions.

In the D1, as shown in Figures~\ref{auc-roc across different datasets} (a) and~\ref{auc-pr across different datasets} (a), the performance of most traditional classifiers such as DT and k-NN deteriorates rapidly when the minority class percentage falls below 25\%. AUC-PR declines more sharply than AUC-ROC, especially for simpler models. TabPFN remains highly robust across all imbalance levels, while CatBoost and SVM show moderate degradation but retain usefulness. In the D2, Figures~\ref{auc-roc across different datasets} (b) and~\ref{auc-pr across different datasets} (b) show that ensemble methods like BaggingRF and RUSBoost perform relatively better under reduced minority conditions, but TabPFN still leads in both AUC metrics. D3 shown in Figures~\ref{auc-roc across different datasets} (c) and~\ref{auc-pr across different datasets} (c), presents a similar trend while boosting models like CatBoost and LightGBM provide some resilience, traditional models degrade significantly. TabPFN again stands out for its consistency under severe imbalance. In D4, Figures~\ref{auc-roc across different datasets} (d) and~\ref{auc-pr across different datasets} (d), all models show a performance drop with increasing imbalance, but TabPFN, CatBoost, and SVM degrade more gracefully. Similarly, in the D5, Figures~\ref{auc-roc across different datasets} (e) and~\ref{auc-pr across different datasets} (e), TabPFN and CatBoost sustain high AUC-ROC and AUC-PR scores down to 1\% minority data, while DT, k-NN, and XGBoost deteriorate rapidly. Overall, systematic reduction experiments confirm that classification becomes more difficult as class imbalance increases, and that advanced models such as TabPFN and boosting ensembles maintain better performance under such conditions.

\begin{figure*}[htbp]
    \centering

    \subfigure[Breast Cancer  Datasets(D1)]{\includegraphics[width=0.45\textwidth]{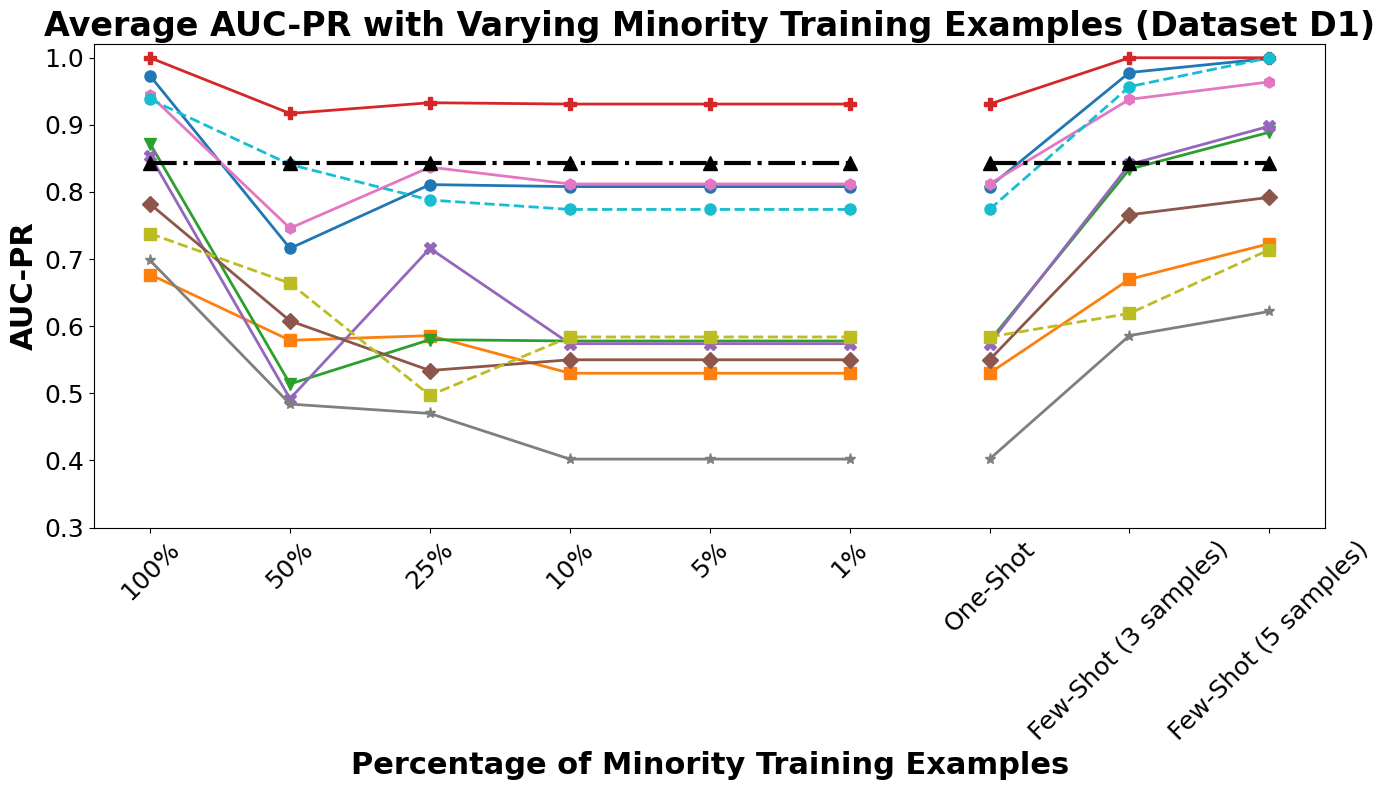}} \hfill
    \subfigure[Pen Local Datasets (D2)]{\includegraphics[width=0.45\textwidth]{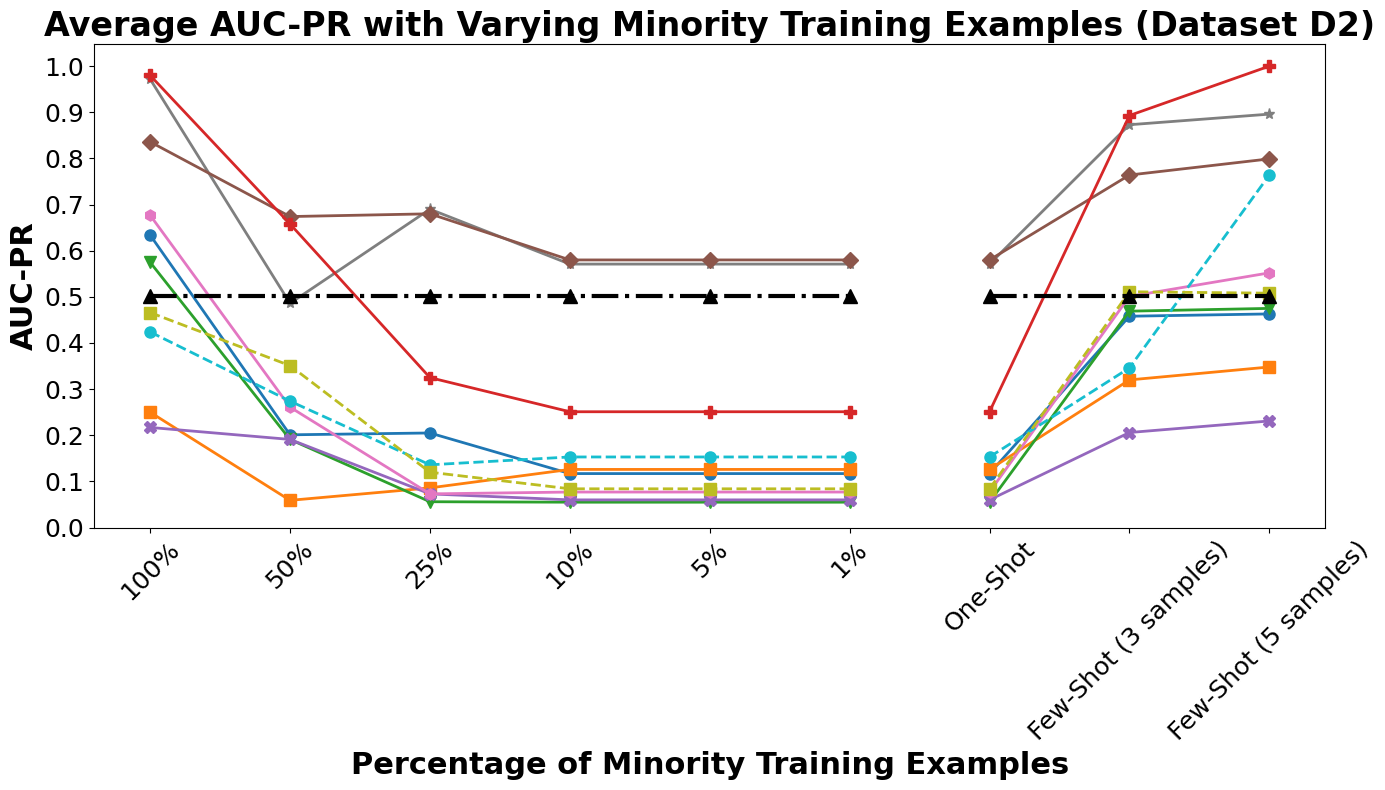}} \\[1ex]

    \subfigure[Pen Global Datasets (D3) ]{\includegraphics[width=0.45\textwidth]{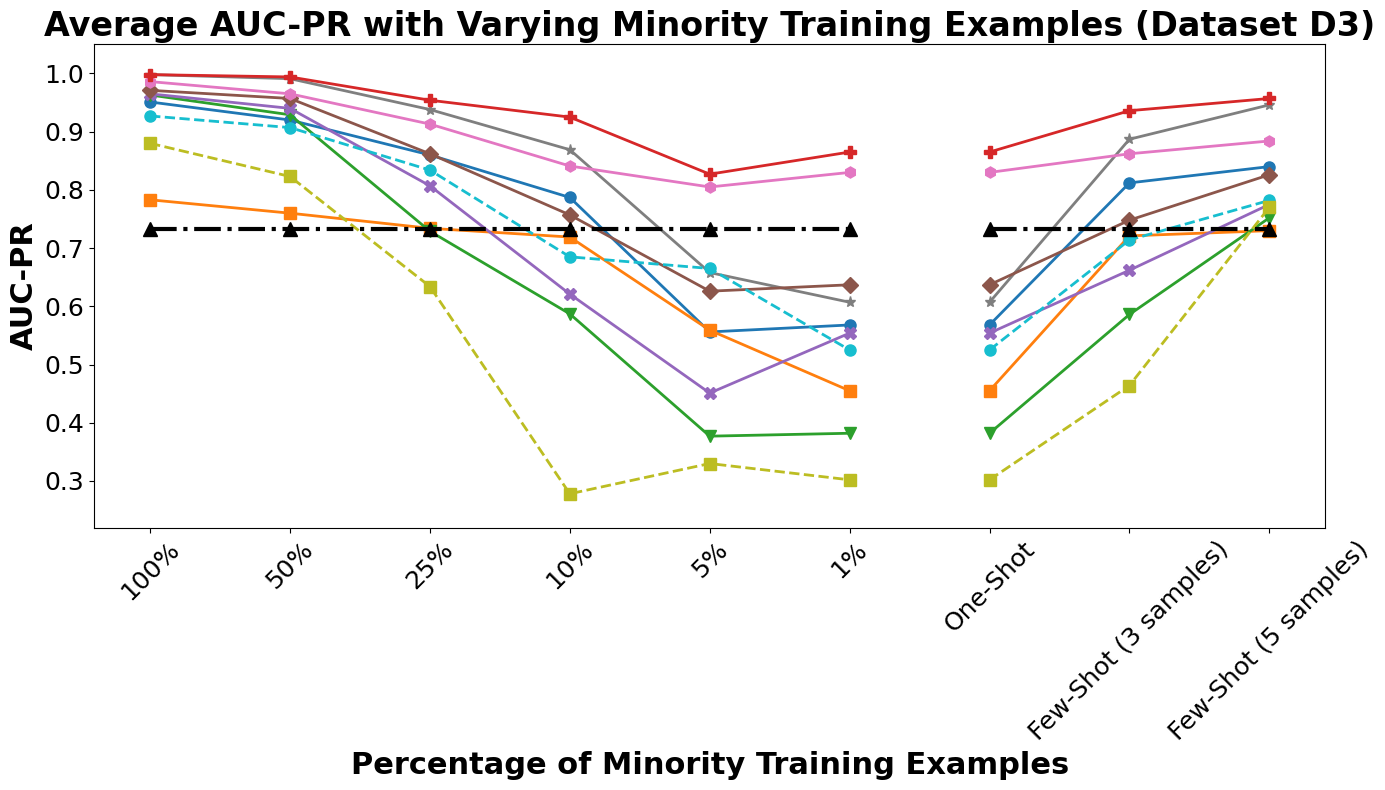}} \hfill
    \subfigure[Letter (D4)]{\includegraphics[width=0.45\textwidth]{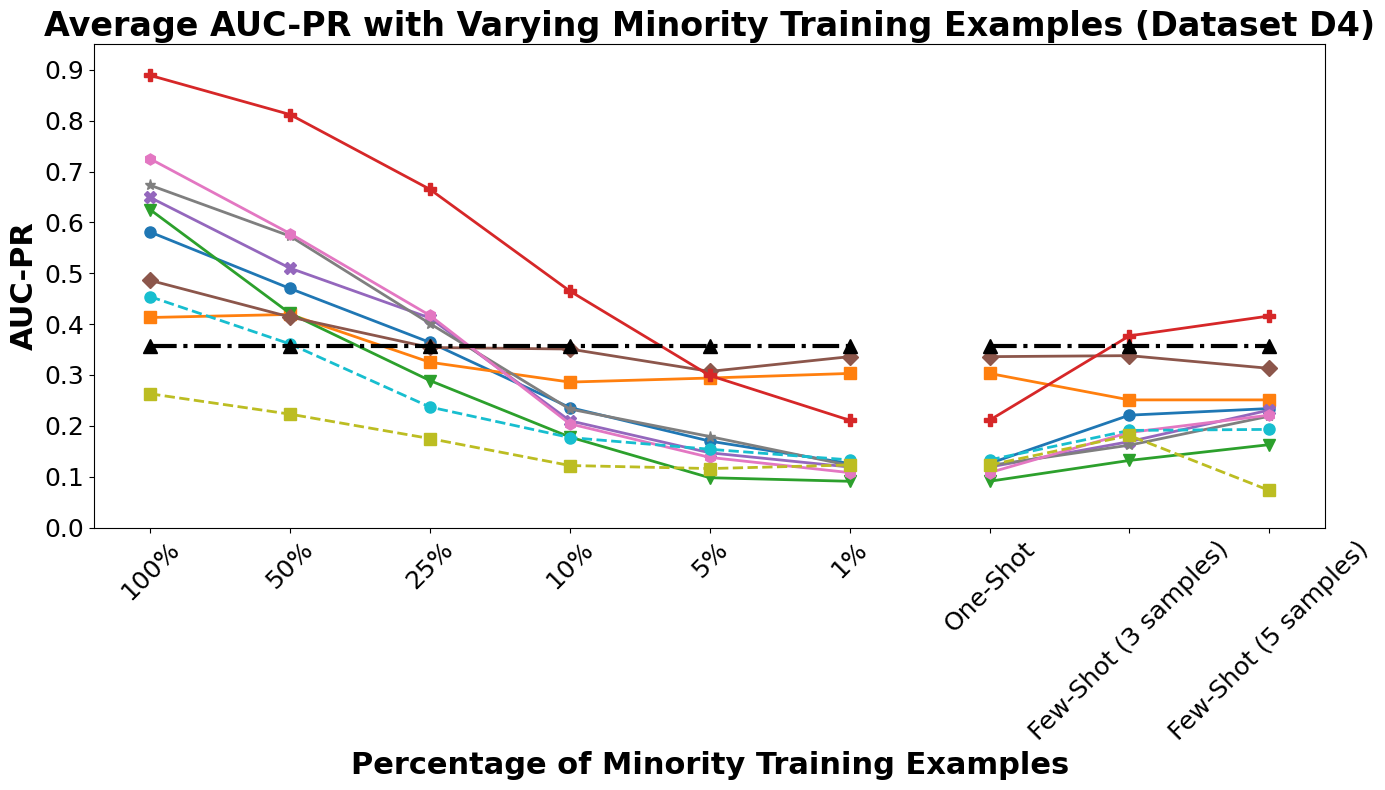}} \\[1ex]

    \subfigure[Annthyroid Datasets (D5)]{\includegraphics[width=0.45\textwidth]{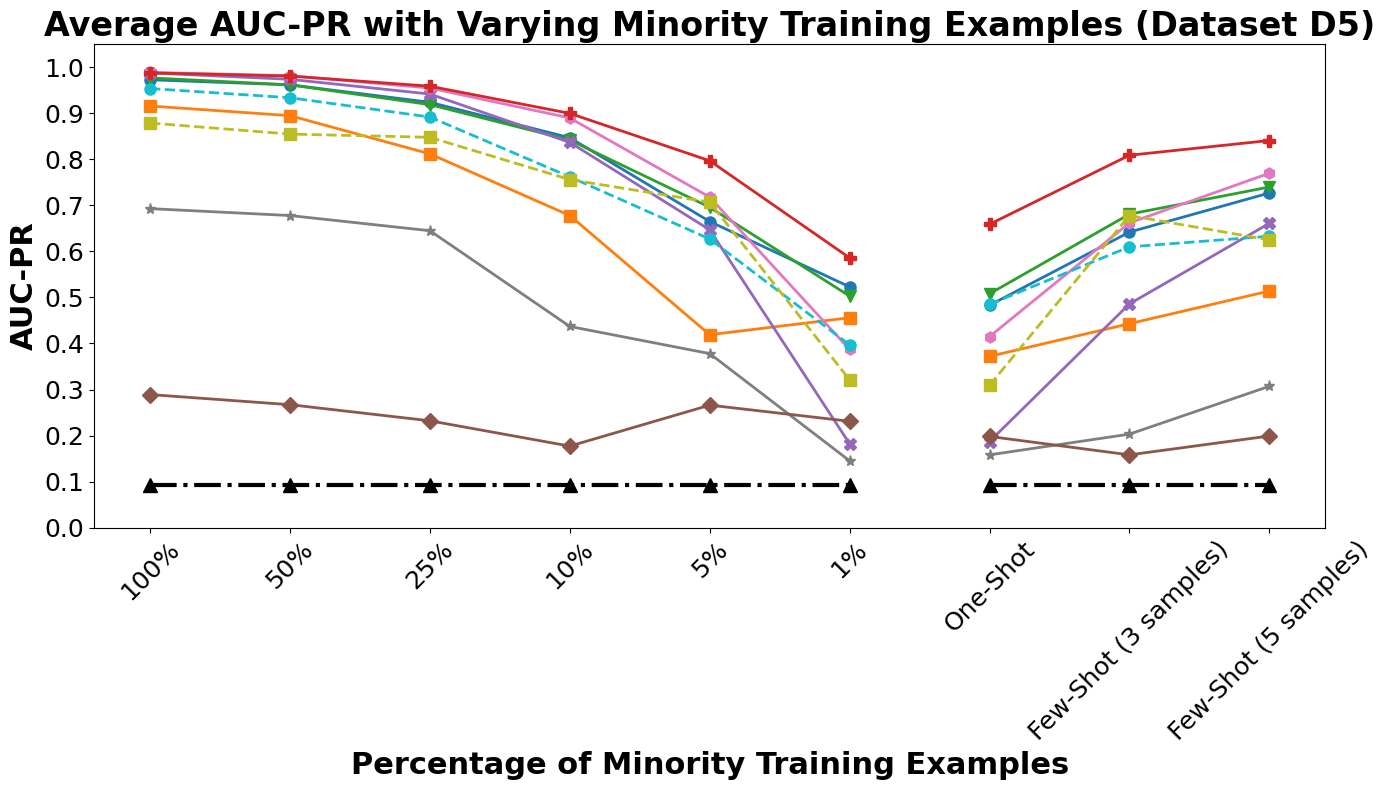}} \\[1ex]

    \caption{Average AUC-PR with varying minority training datapoints of different datasets.}
    \label{auc-pr across different datasets}
\end{figure*}

\subsection{Robustness under One-shot and Few-shot}

To evaluate classifier robustness under extreme data scarcity, we explored one-shot and few-shot scenarios, where the number of minority class samples is limited to 1, 3, and 5 instances. These conditions simulate real-world cases where collecting minority data is expensive or infeasible.

Our results show that even a small increase from one to three or five minority samples can significantly improve performance for most classifiers. However, the ability to generalize under such limited data remains highly model-dependent. While OCC models inherently limited by their reliance on majority class only, demonstrated surprising robustness on certain datasets, highlighting their utility as a fallback when no minority data is accessible.

Among all models, TabPFN maintained the strongest performance across one-shot, few-shot, and OCC settings, reflecting its robustness in extreme imbalance conditions. Ensemble methods such as BaggingRF and RUSBoost showed moderate improvements but did not match the generalization capacity of TabPFN or CatBoost. These findings emphasize that robustness is key when working under data scarcity, and advanced probabilistic or deep tabular models offer promising solutions.

The one-shot and few-shot experiments (1, 3, and 5 minority examples) located at the right side of each subfigure in Figures~\ref{auc-roc across different datasets} and~\ref{auc-pr across different datasets} highlight the behaviour of classifiers under extremely low data regimes. For D1, most models, including DT and RF, perform poorly in the one-shot setting, with AUC-PR falling close to random. However, TabPFN delivers strong and stable performance, followed by moderate recovery by CatBoost and SVM in the 3 and 5-shot conditions. In the D2, TabPFN again dominates the one-shot scenario with high AUC scores, and ensemble methods like BaggingRF and RUSBoost improve as more minority samples are added. The D3 reinforces this trend: one-shot performance is unreliable for most models, but TabPFN and SVM show promising results, and boosting methods catch up in few-shot settings. In the D4, one-shot AUC-PR is very low for most classifiers except TabPFN, which maintains separation between classes even with one minority sample. CatBoost and SVM again show improvement at 3 and 5 samples. For the D5, TabPFN delivers the best performance under all low-data settings, while others including RF, DT, and k-NN perform inconsistently or collapse entirely in one-shot settings. These findings clearly demonstrate that most classifiers fail under extreme minority scarcity, but models such as TabPFN, CatBoost, and SVM are capable of effective learning even with only a few minority instances, validating their suitability for real-world, low-resource classification tasks.

\subsection{Synthetic Decision Boundary Generation}

\begin{figure*}[htbp]
    \centering

    \subfigure[Linear High Separation]{\includegraphics[width=0.45\textwidth]{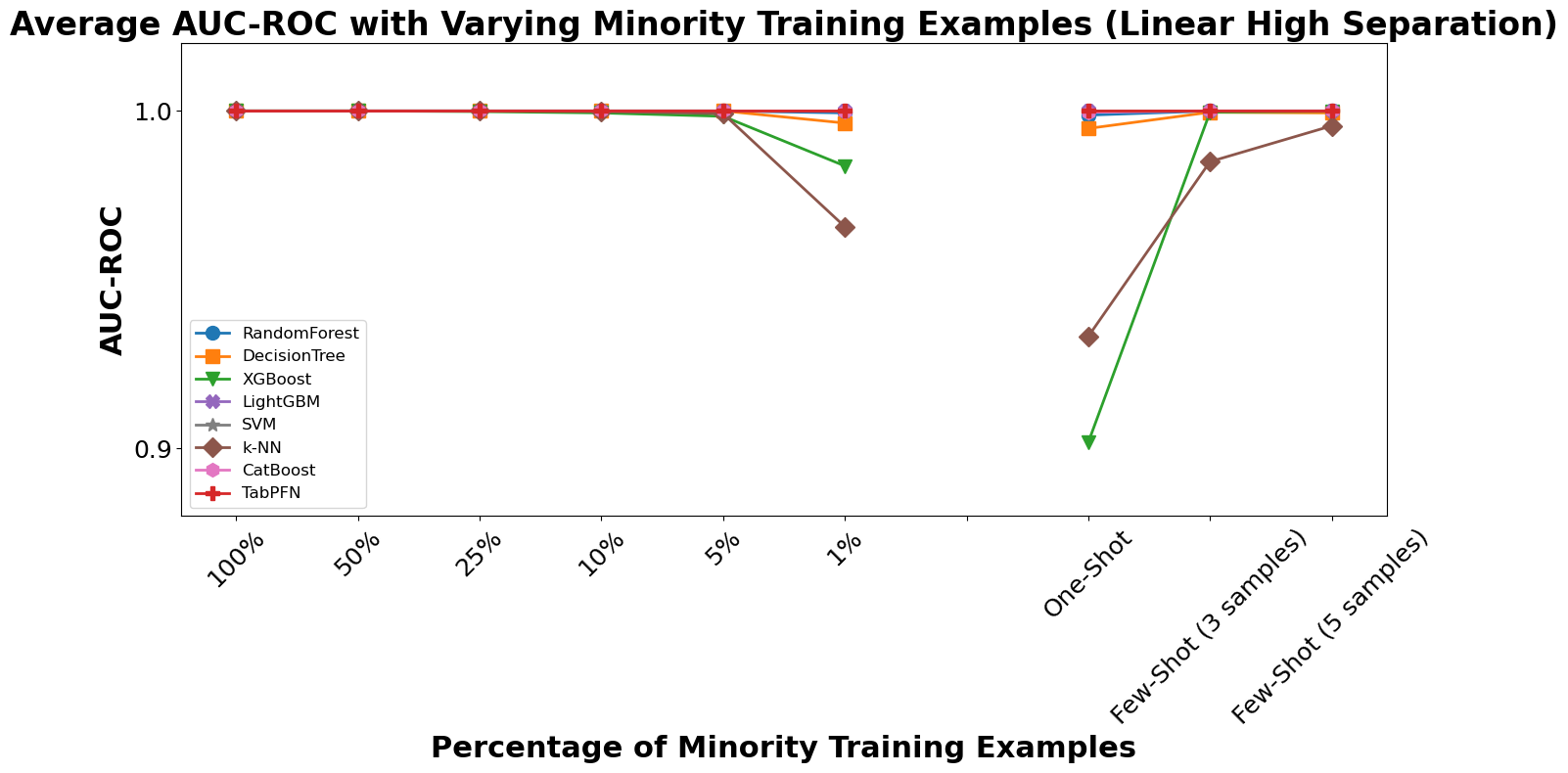}} \hfill
    \subfigure[Moderate Non Linearity]{\includegraphics[width=0.45\textwidth]{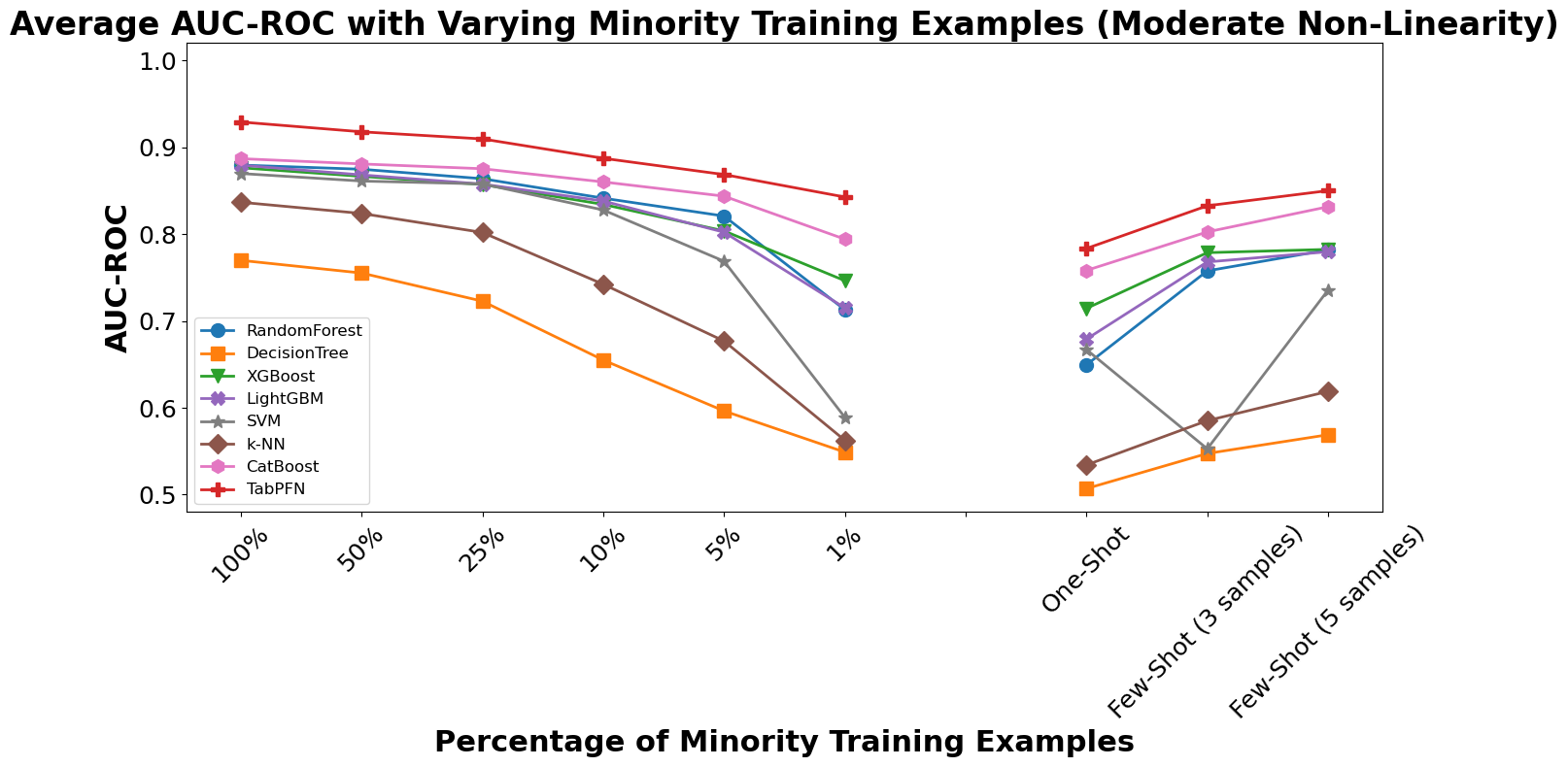}} \\[1ex]

    \subfigure[Non-linear Redundancy+PCA]{\includegraphics[width=0.45\textwidth]{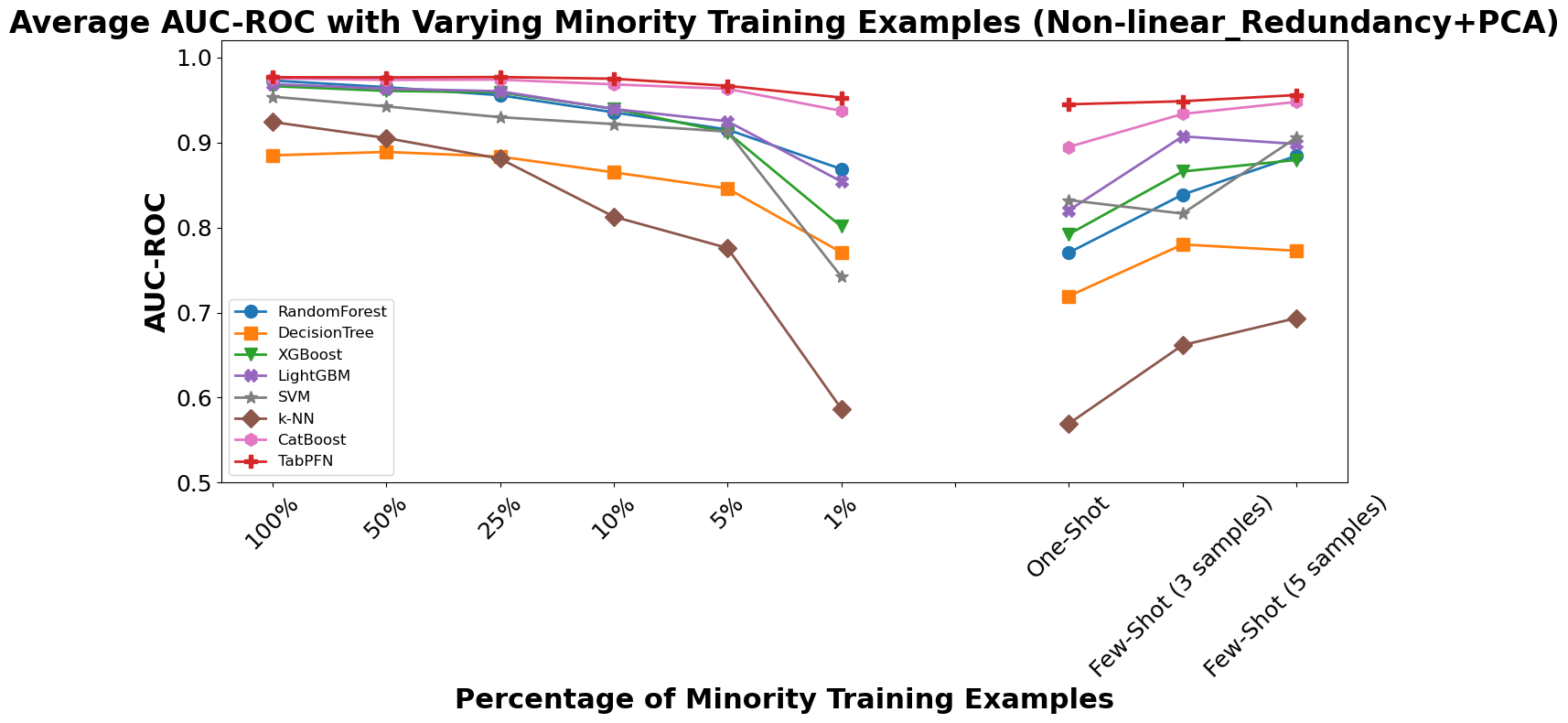}} \hfill
    \subfigure[Non-linear gassian quantiles]{\includegraphics[width=0.45\textwidth]{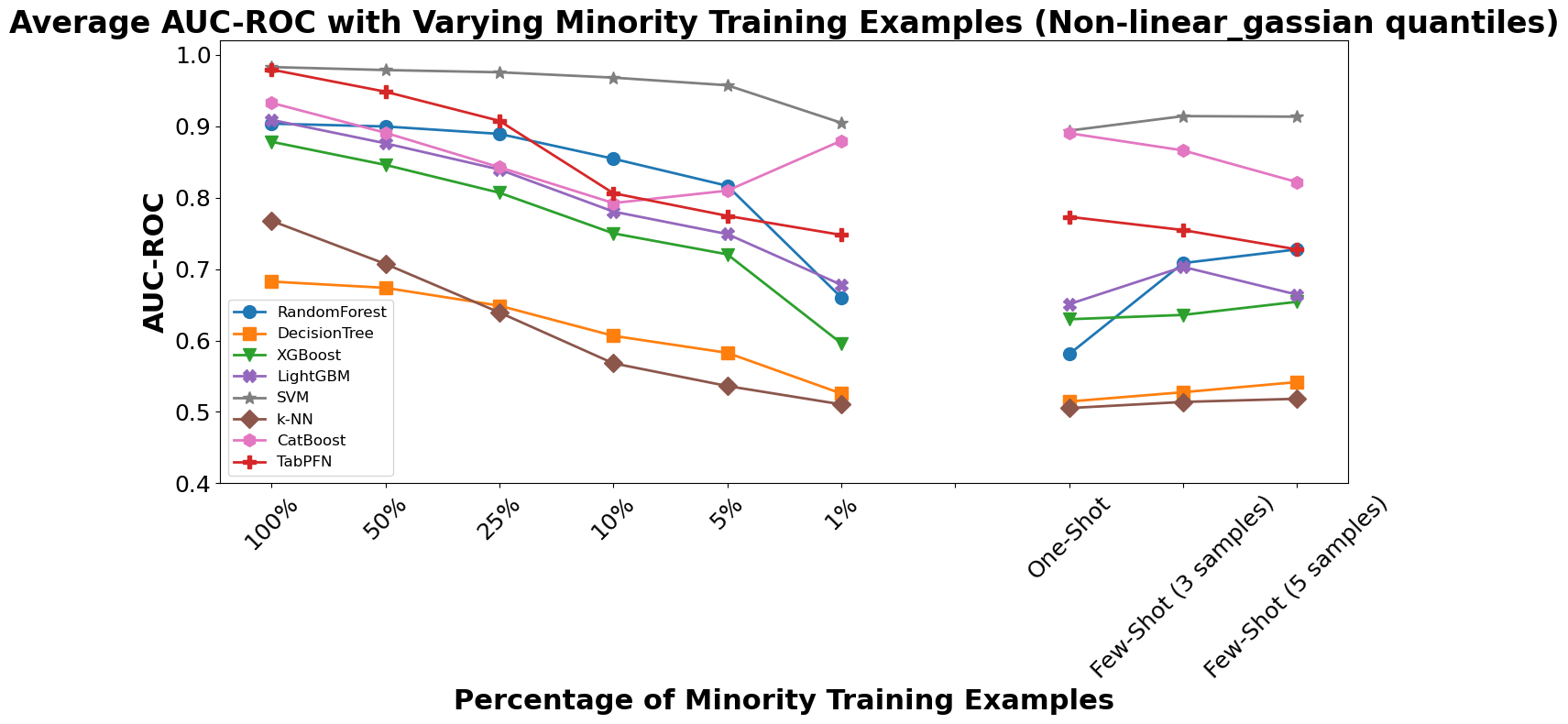}} \\[1ex]

    \subfigure[Hard Non-linear XOR]{\includegraphics[width=0.45\textwidth]{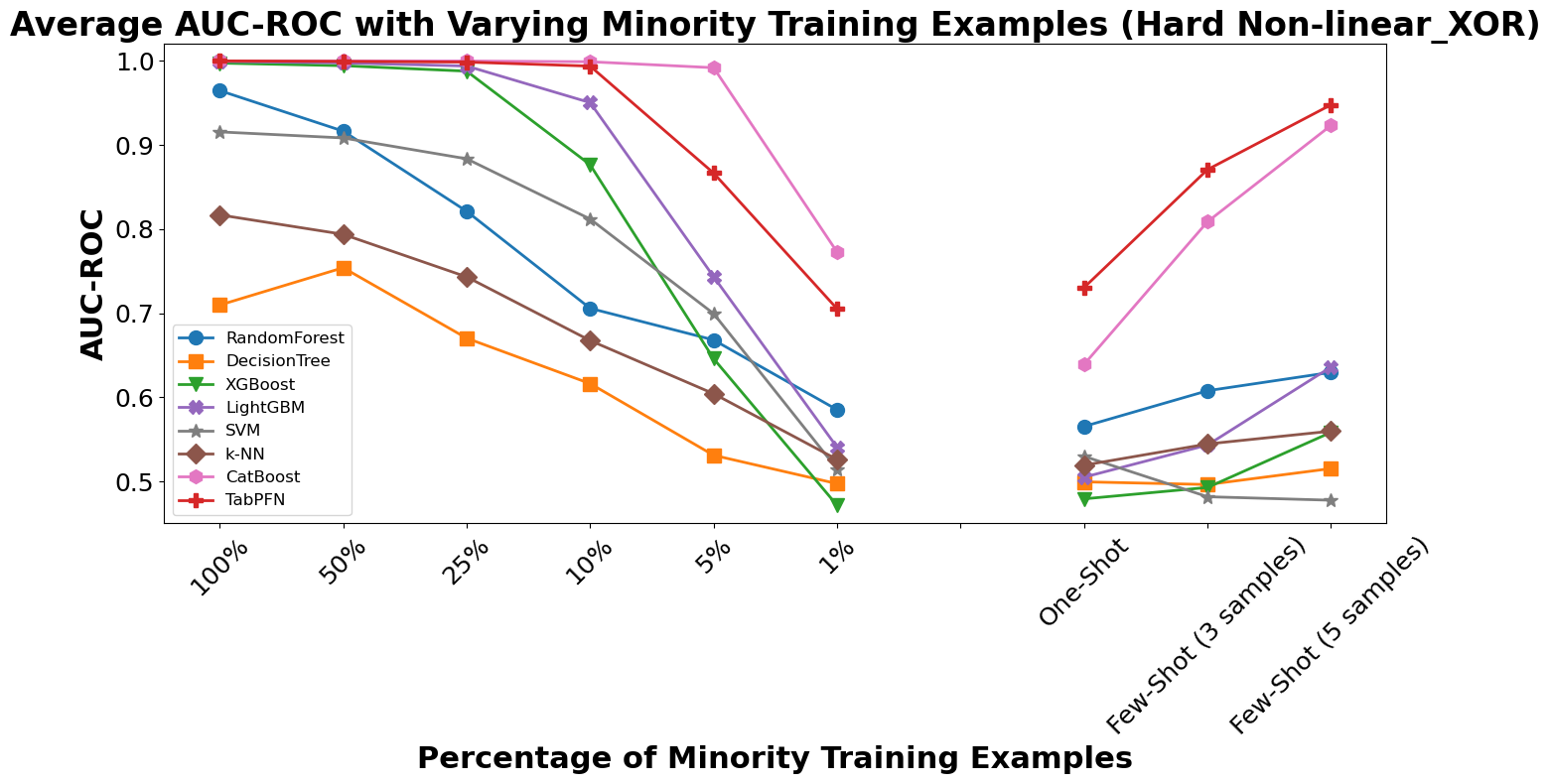}} \\[1ex]

    \caption{Average AUC-ROC of classification models across five synthetic datasets, each representing a distinct decision boundary}
    \label{auc-ROC datasets}
\end{figure*}

\noindent
Figure~\ref{auc-ROC datasets} presents the AUC-ROC performance of classification models across five synthetic datasets, each constructed to simulate distinct decision boundary complexities. These results support the goal of identifying the minimum imbalance ratio that enables reliable classification and generalization. The datasets vary from simple linearly separable structures to highly complex non-linear geometries, providing a comprehensive evaluation of classifier robustness under progressively reduced minority class sizes. Figure~\ref{auc-ROC datasets}~(a) represents results of linearly separable dataset with high class separation. Here, most classifiers achieve good AUC-ROC scores across all imbalance levels, including one-shot and few-shot scenarios, demonstrating that in such ideal geometric conditions, reliable classification is still achievable even with extreme class imbalance. In Figure~\ref{auc-ROC datasets}~(b), moderate non-linearity is introduced, leading to visible performance degradation in simpler models such as k-NN and Decision Tree as imbalance increases. However, robust models like CatBoost and TabPFN retain relatively strong performance, suggesting their capacity to generalize even when minority data is scarce. Figure~\ref{auc-ROC datasets}~(c) adds feature redundancy and non-linear transformations followed by PCA, introducing additional complexity and reducing feature distinctiveness. As the imbalance grows, classifiers experience sharper declines in AUC-ROC, with only a few models maintaining acceptable performance under few-shot conditions. Figure~\ref{auc-ROC datasets}~(d) presents the results of non-linear Gaussian quantile distributions with overlapping class boundaries. The results show a consistent performance drop across all models as minority class size decreases, revealing that generalization becomes difficult under noisy, non-separable conditions. Finally, Figure~\ref{auc-ROC datasets}~(e), the minority class follows a highly non-linear decision boundary. In this scenario, nearly all classifiers exhibit sharp performance declines under increasing imbalance. Only the most advanced models such as TabPFN and CatBoost demonstrate moderate resilience in the few-shot regime.

\begin{figure*}[htbp]
    \centering

    \subfigure[Linear High Separation]{\includegraphics[width=0.45\textwidth]{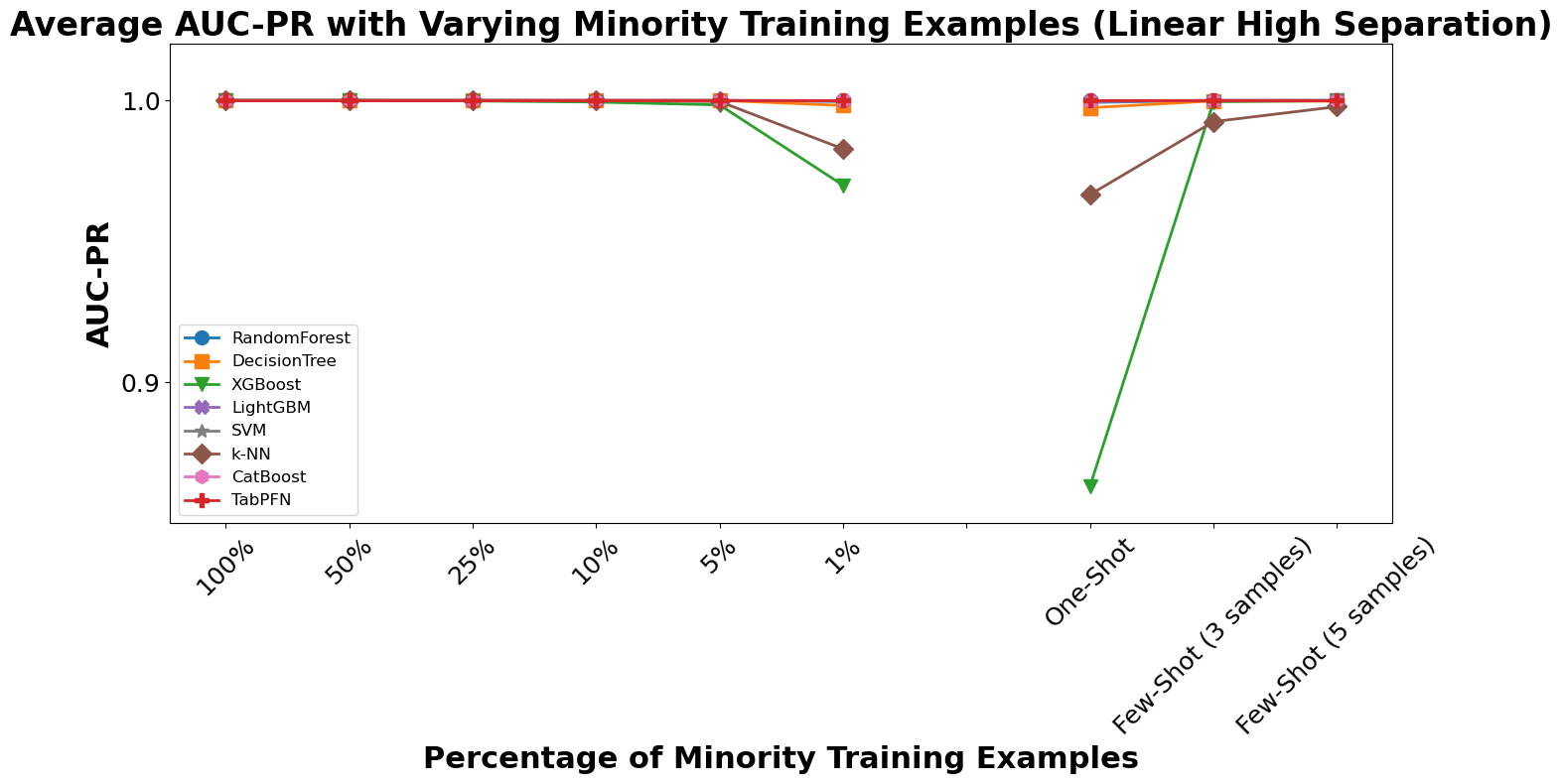}} \hfill
    \subfigure[Moderate Non Linearity]{\includegraphics[width=0.45\textwidth]{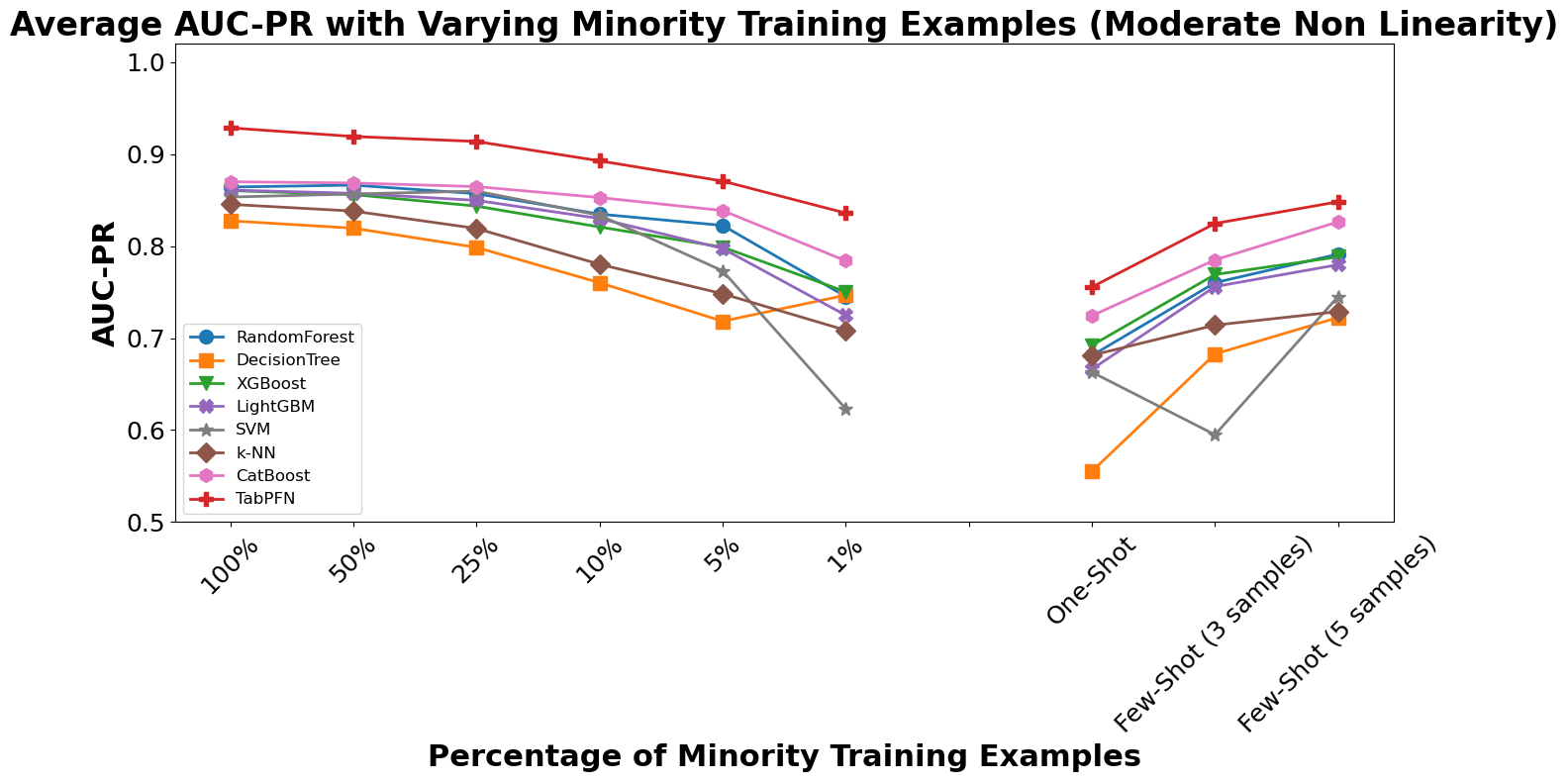}} \\[1ex]

    \subfigure[Non-linear Redundancy+PCA]{\includegraphics[width=0.45\textwidth]{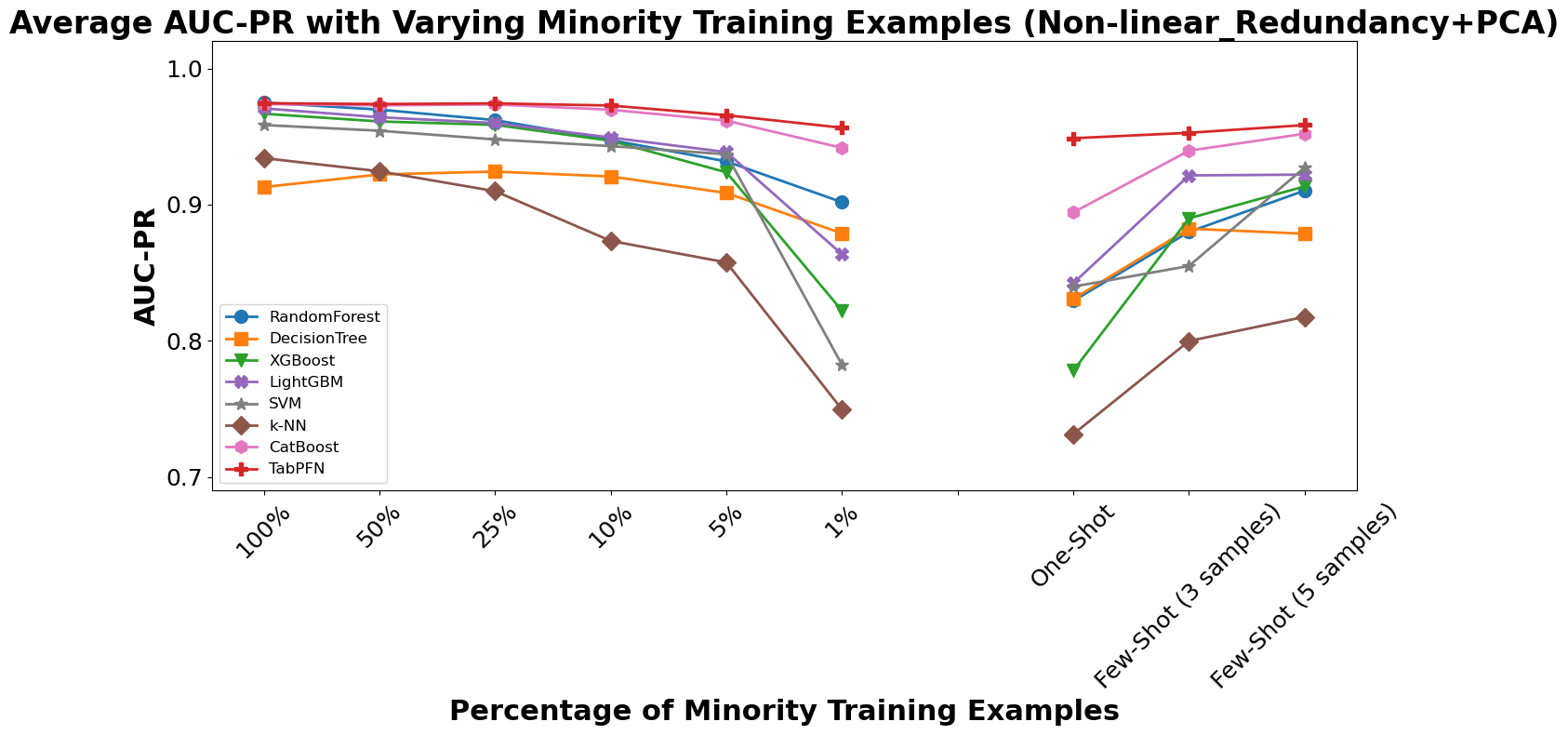}} \hfill
    \subfigure[Non-linear gassian quantiles]{\includegraphics[width=0.45\textwidth]{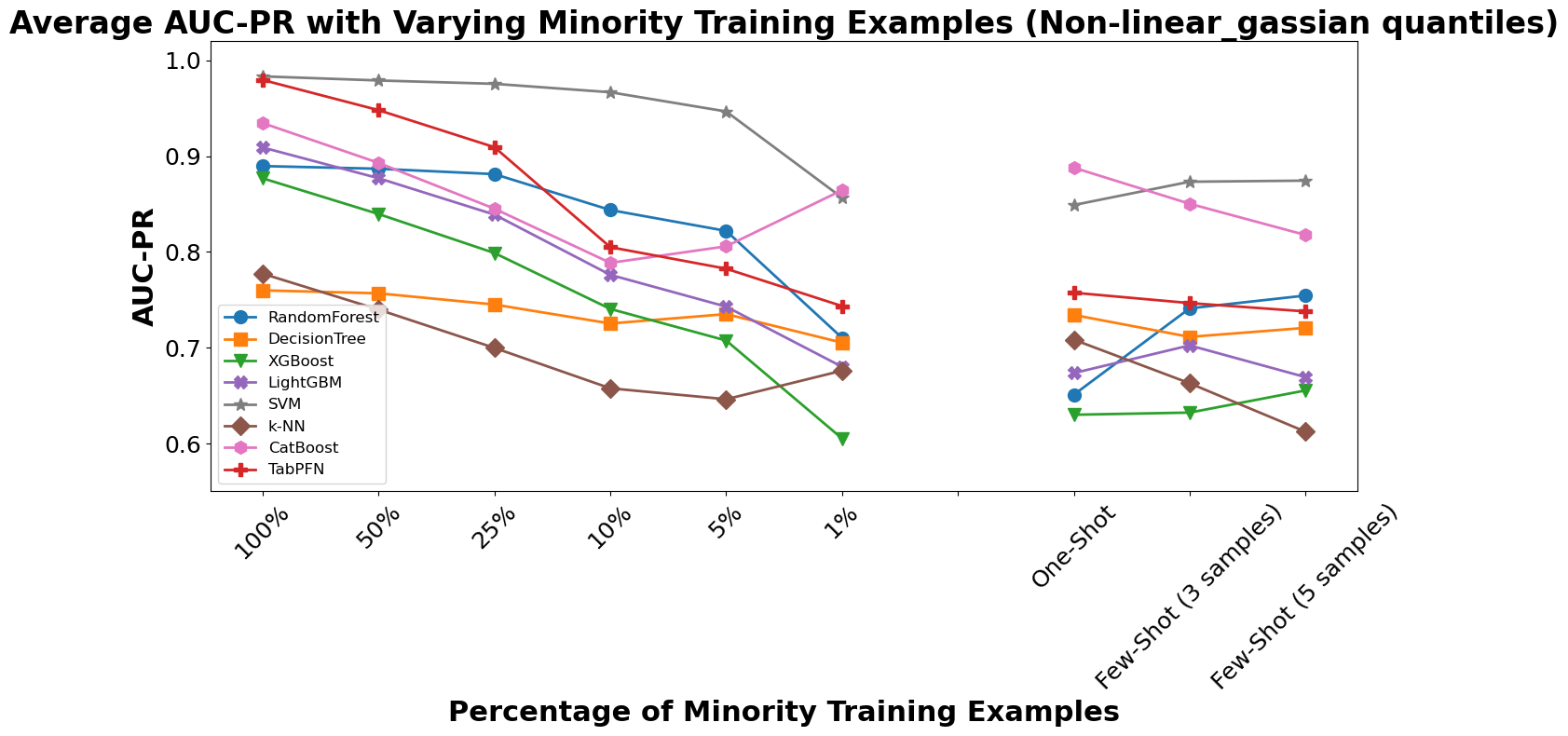}} \\[1ex]

    \subfigure[Hard Non-linear XOR]{\includegraphics[width=0.45\textwidth]{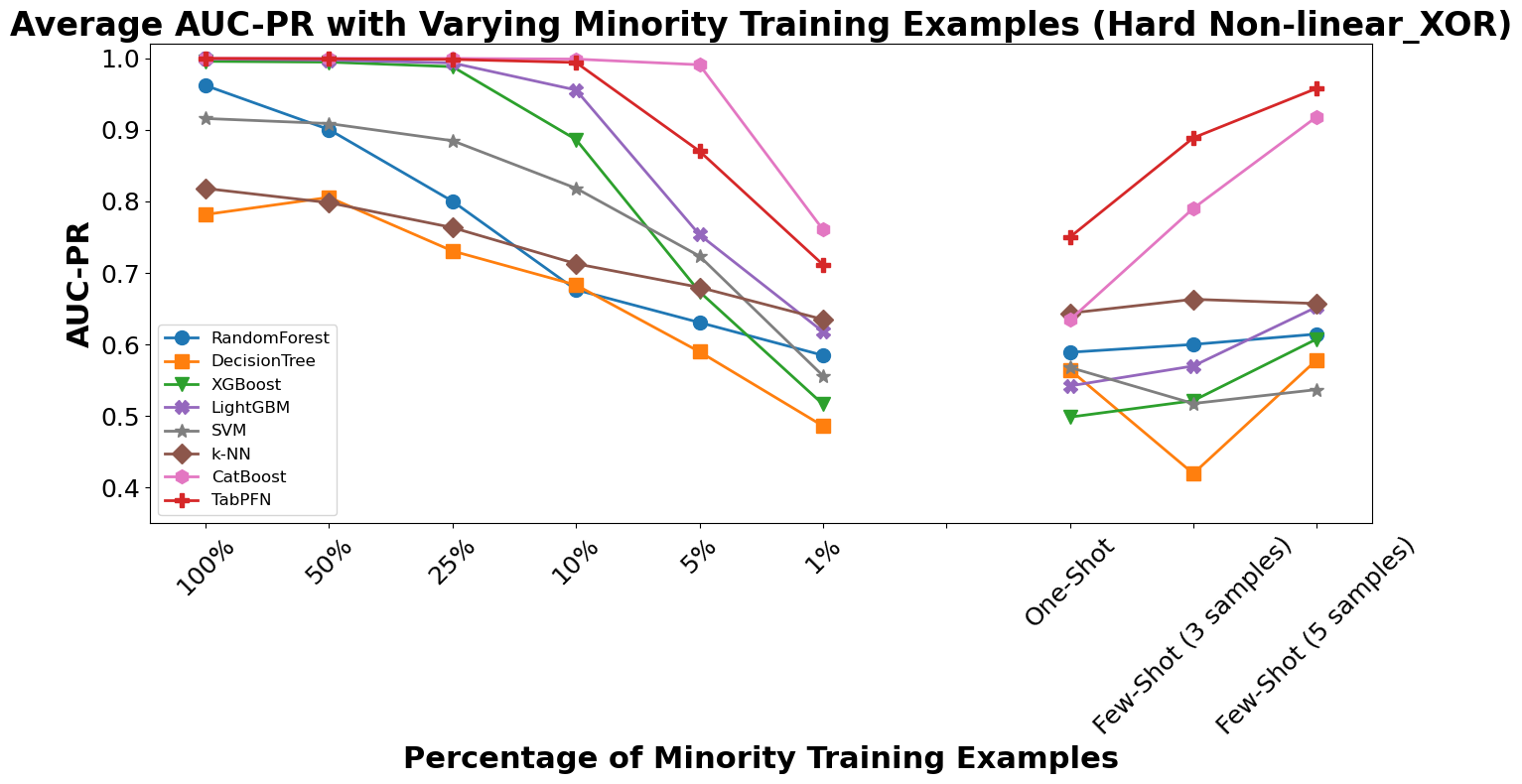}} \\[1ex]

    \caption{Average AUC-PR of classification models across five synthetic datasets, each representing a distinct decision boundary}
    \label{auc-pr datasets}
\end{figure*}

Figure~\ref{auc-pr datasets} illustrates the AUC-PR performance of various binary classifiers across five synthetic datasets, each crafted to exhibit unique structural and geometric complexity. The results serve to pinpoint the minimum viable imbalance ratio where models can still generalize effectively, a critical factor in high-stakes domains with severely imbalanced classes. Figure~\ref{auc-pr datasets}~(a) presents the performance on a linearly separable dataset with high class distinction. As expected, nearly all models maintain high AUC-PR scores across the full range of imbalance levels, including the extreme one-shot and few-shot scenarios. This indicates that when the data geometry is simple and the class boundaries are well defined, precision-recall performance remains resilient even under severe data scarcity. Figure~\ref{auc-pr datasets}~(b), which reflects moderate non-linearity results, shows a more prominent drop in AUC-PR as imbalance increases. Classical model such as DT and k-NN degrade rapidly, while more sophisticated models like TabPFN and CatBoost demonstrate relatively stable performance, highlighting their ability to maintain precision and recall in more complex data environments. In Figure~\ref{auc-pr datasets}~(c), the dataset includes feature redundancy and non-linear transformations followed by PCA, adding complexity to the feature space, which leads to sharper declines in AUC-PR, particularly for models that are sensitive to irrelevant features or low sample density. Only a few robust models manage to retain acceptable AUC-PR levels under few-shot scenarios. Figure~\ref{auc-pr datasets}~(d) features Gaussian quantile distributions with overlapping boundaries, creating a challenging classification setting. The drop in AUC-PR across all models emphasizes the difficulty of maintaining high precision when the minority class is both sparse and indistinct, revealing the limits of generalization in noisy, and overlapping data. Finally, Figure~\ref{auc-pr datasets}~(e) represents XOR configuration, characterized by a hard non-linear decision boundary. Most classifiers experience significant performance degradation, particularly in the few-shot regime. The consistent superiority of TabPFN and CatBoost in this setting underlines their advanced capacity to handle high-complexity distributions, even when training examples are minimal.

To summarize, these experiments confirm that classifier performance under class imbalance is dependent on data complexity and geometry. The results revealed that performance decreases as the decision boundary becomes more complicated. Reliable classification can be achieved at extremely low imbalance ratios only when the underlying structure is simple or when robust learning algorithms are applied.

\begin{figure*}[htbp]
    \centering
    \subfigure[100\% minority points, Breast Cancer Datasets (D1), DT]{
        \includegraphics[width=0.45\textwidth]{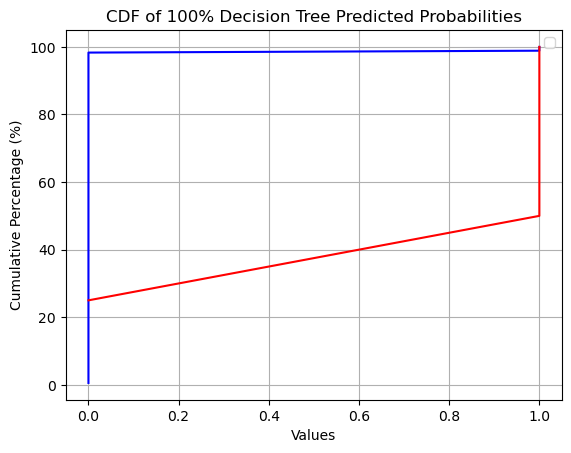}
    }
    \hfill
    \subfigure[100\% minority points, Breast Cancer Datasets (D1), CatBoost]{
        \includegraphics[width=0.45\textwidth]{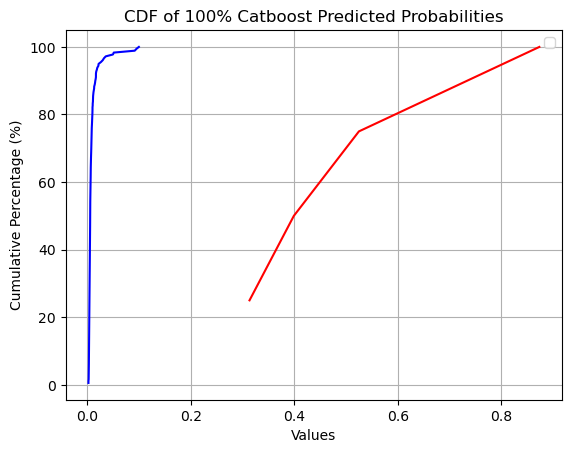}
    }
    \\
    \subfigure[100\% minority points, Breast Cancer Datasets (D1), TabPFN]{
        \includegraphics[width=0.45\textwidth]{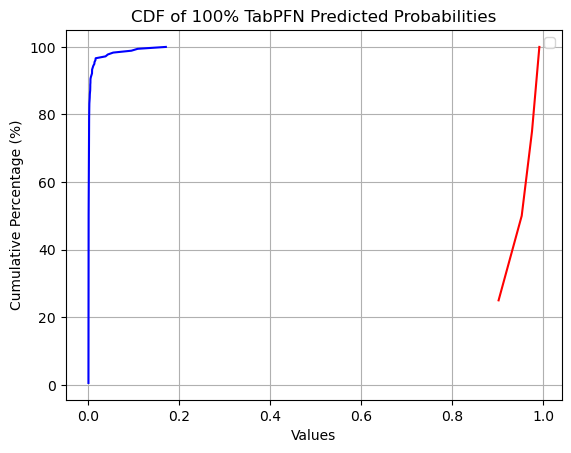}
    }
    \hfill
    \subfigure[100\% minority points, Pen Local Datasets (D2), DT]{
        \includegraphics[width=0.45\textwidth]{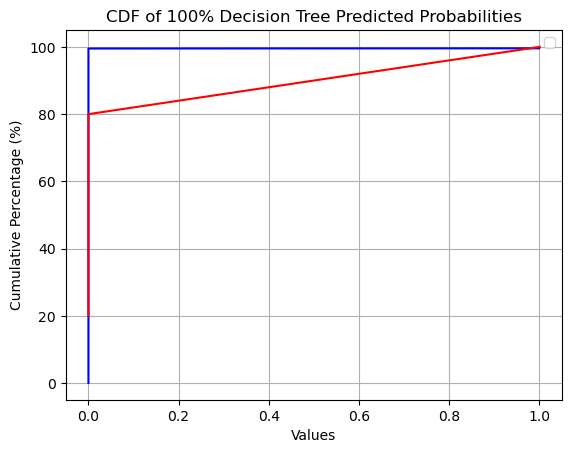}
    }
    \\
    \subfigure[100\% minority points, Pen Local Datasets (D2), CatBoost]{
        \includegraphics[width=0.45\textwidth]{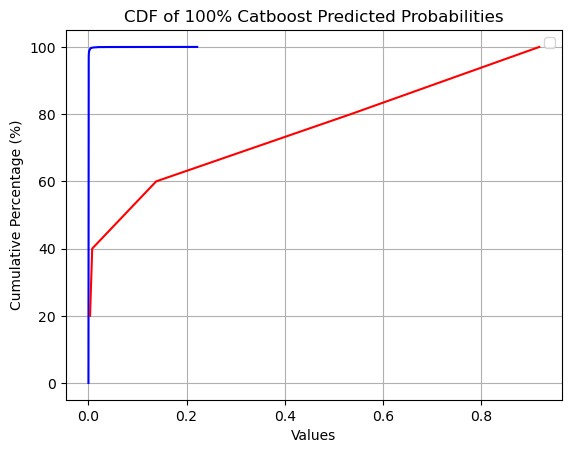}
    }
    \hfill
    \subfigure[100\% minority points, Pen Local Datasets (D2), TabPFN]{
        \includegraphics[width=0.45\textwidth]{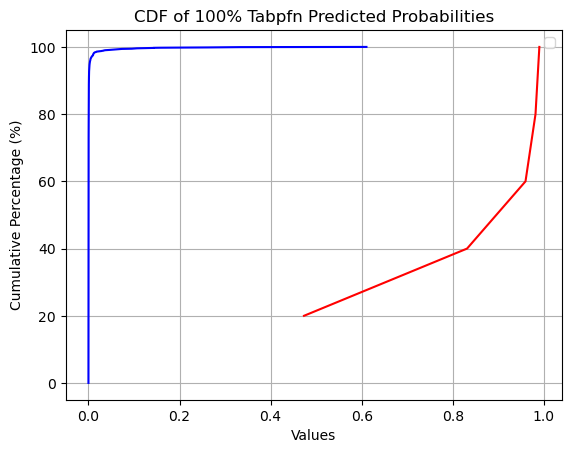}
    }

    \caption{Cumulative percentage vs. predicted probability plots for majority (blue) and minority class datapoints (red) generated by various classification models (DT, CatBoost, and TabPFN) on the datasets (Breast cancer and Pen local datasets). The X-axis represents the model’s output score, while the Y-axis shows the cumulative distribution percentage of samples. These plots illustrate the model's separation ability between classes; sharper divergence between red and blue curves indicates better discriminative performance.}
    \label{decision_vis}
\end{figure*}

\begin{figure*}[htbp]
    \centering
    \subfigure[100\% minority points, Pen Global Datasets (D3), DT]{
        \includegraphics[width=0.45\textwidth]{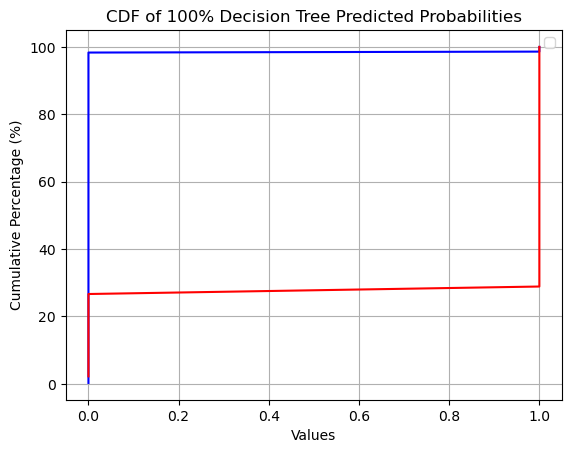}
    }
    \hfill
    \subfigure[100\% minority points, Pen Global Datasets (D3), CatBoost]{
        \includegraphics[width=0.45\textwidth]{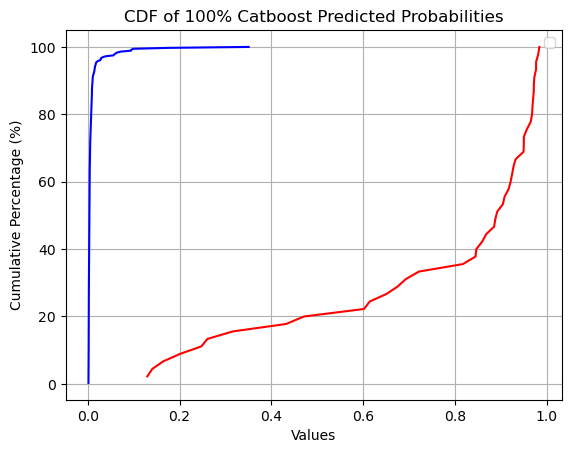}
    }
    \\
    \subfigure[100\% minority points, Pen Global Datasets (D3), TabPFN]{
        \includegraphics[width=0.45\textwidth]{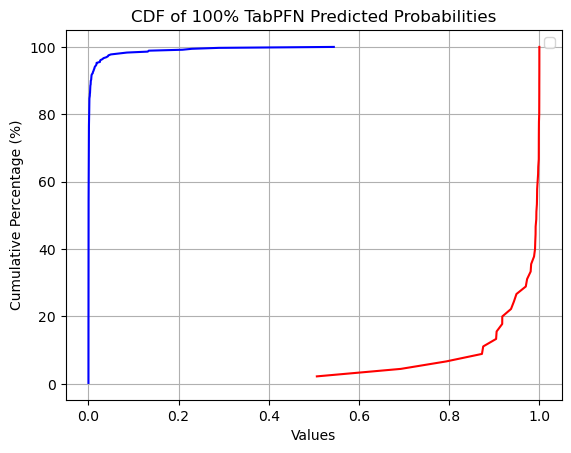}
    }
    \hfill
    \subfigure[100\% minority points, Letter Datasets (D4), DT]{
        \includegraphics[width=0.45\textwidth]{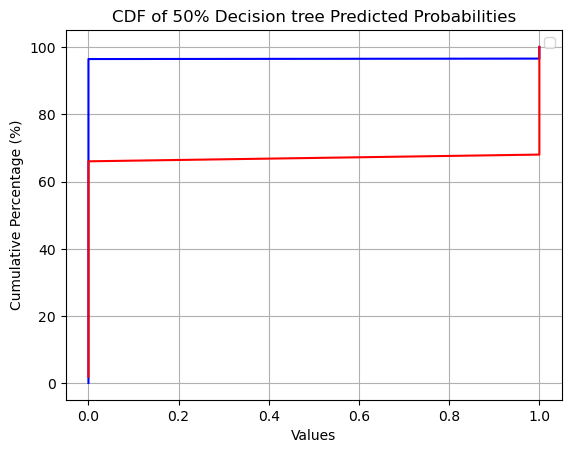}
    }
    \\
    \subfigure[100\% minority points, Letter Datasets (D4), CatBoost]{
        \includegraphics[width=0.45\textwidth]{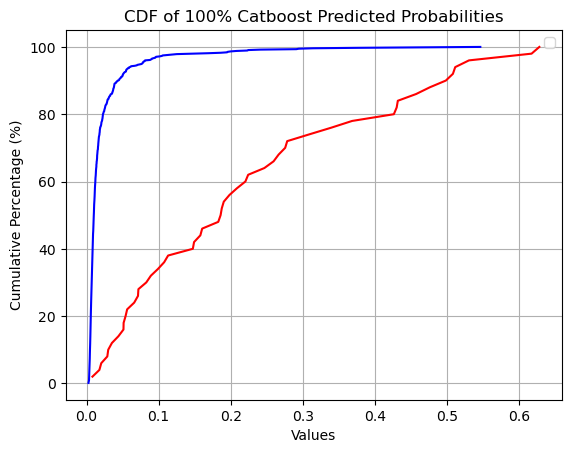}
    }
    \hfill
    \subfigure[100\% minority points, Letter Datasets (D4), TabPFN]{
        \includegraphics[width=0.45\textwidth]{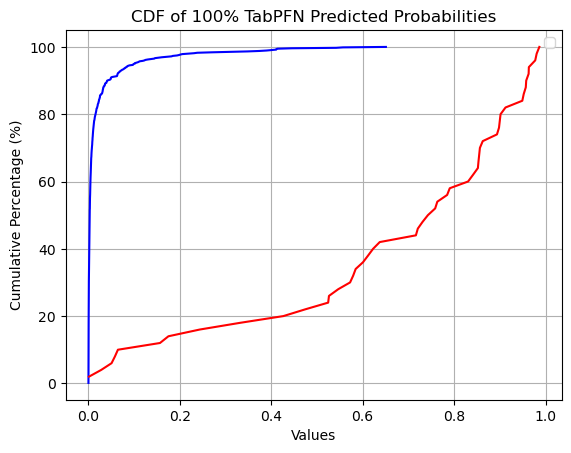}
    }

    \caption{Cumulative percentage vs. predicted probability plots for majority (blue) and minority class datapoints (red) generated by various classification models (DT, CatBoost, and TabPFN) on the datasets (Pen Global and Letter datasets). The X-axis represents the model’s output score, while the Y-axis shows the cumulative distribution percentage of samples. These plots illustrate the model's separation ability between classes; sharper divergence between red and blue curves indicates better discriminative performance.}
    \label{decision_vis1}
\end{figure*}

\begin{figure}[ht]
    \centering
    \subfigure[100\% minority points, Annthyroid Datasets (D5), DT]{
        \includegraphics[width=0.45\textwidth]{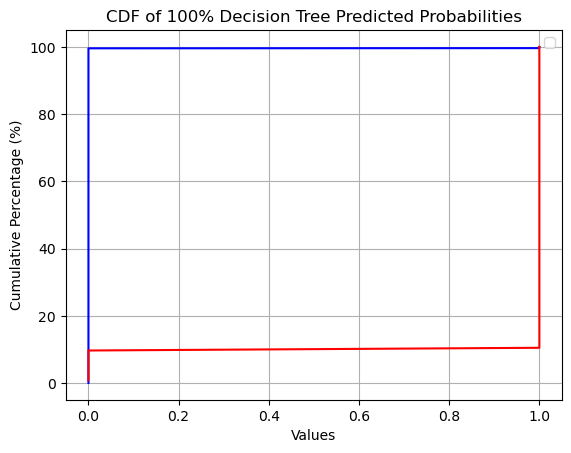}
    }
    \hfill
    \subfigure[100\% minority points, Annthyroid Datasets (D5), CatBoost]{
        \includegraphics[width=0.45\textwidth]{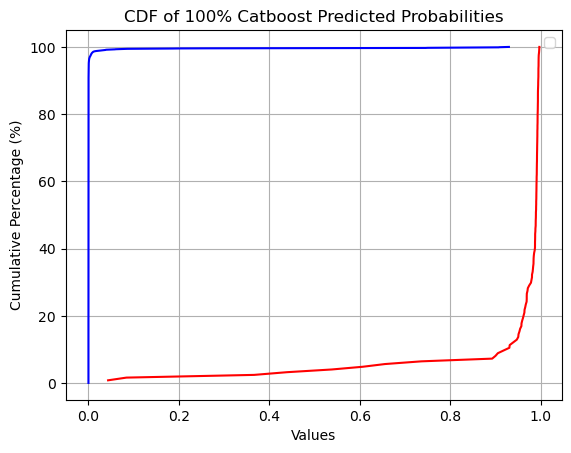}
    }
    \\
    \subfigure[100\% minority points, Annthyroid Datasets (D5), TabPFN]{
        \includegraphics[width=0.45\textwidth]{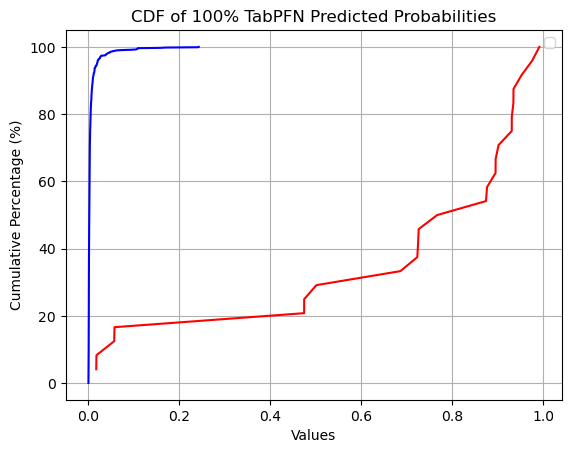}
    }

    \caption{Cumulative percentage vs. predicted probability plots for majority (blue) and minority class datapoints (red) generated by various classification models (DT, CatBoost, and TabPFN) on the datasets (Annthyroid datasets). The X-axis represents the model’s output score, while the Y-axis shows the cumulative distribution percentage of samples. These plots illustrate the model's separation ability between classes; sharper divergence between red and blue curves indicates better discriminative performance. Due to space constraints, a limited set of representative visualizations is included to illustrate key trends across models and datasets.}
    \label{decision_vis2}
\end{figure}

\subsection{Analysis of prediction score of different classification models}

Figures~\ref{decision_vis},\ref{decision_vis1}, and \ref{decision_vis2} present Cumulative Distribution Function (CDF) plots of predicted probabilities for majority and minority classes, generated by three classifiers i.e., DT, CatBoost, and TabPFN on five representative datasets. These plots visualize the model’s ability to distinguish between classes by comparing how confidently each model assigns prediction scores. The blue curves represent the cumulative percentage of majority class instances with respect to their predicted scores, while the red curves show the distribution for the minority class. A sharp separation between these two curves indicates strong discriminative performance, whereas overlapping curves suggest weak class separation.

In the Breast Cancer dataset (D1), shown in Figures~\ref{decision_vis} (a–c), the DT model (a) shows limited separation between classes, as the red curve overlaps significantly with the blue curve. CatBoost (b) performs better, pushing the red curve farther to the right, but TabPFN (c) achieves the clear separation, with the red curve spreading across high-probability regions while the blue curve remains tightly concentrated at low values. Similar trends are observed in the Pen Local dataset (D2), where DT again exhibits poor separation in Figure~\ref{decision_vis} (d), CatBoost offers moderate class distinction (e), and TabPFN (f) sharply distinguishes the two distributions, underscoring its robustness in capturing decision boundaries under imbalance.

Figure~\ref{decision_vis1} further validates these findings using Pen Global (D3) and Letter (D4) datasets. In D3, the DT model (a) performs poorly, as shown by overlapping CDFs, while CatBoost (b) displays better separation. TabPFN (c) once again excels, with steep and well-separated curves, indicating high model confidence. In the Letter dataset (D4), both CatBoost (e) and TabPFN (f) provide better separation compared to DT (d), although the separation is less pronounced than in simpler datasets suggesting increased difficulty due to class overlap and feature complexity.

Finally, Figure~\ref{decision_vis2} illustrates CDF plots for the Annthyroid dataset (D5). DT (a) and CatBoost (b) produce partially overlapping curves, which hints at ambiguity in their class assignment. However, TabPFN (c) maintains strong separation, reflecting its consistent confidence and superior ability to model decision boundaries. The red curve, corresponding to minority predictions, is well-dispersed toward high prediction values, while the blue curve remains steep and left-skewed, indicating low scores for normal instances.

Overall, the CDF visualizations demonstrate that TabPFN outperforms traditional classifiers by maintaining clear and confident decision boundaries across various datasets. Its ability to generate well-separated distributions for majority and minority classes even under imbalanced conditions reaffirms its suitability for real-world anomaly and minority class detection problems.

\section{Conclusion}
To conclude, our study systematically evaluated the robustness of binary classifiers under class imbalance without explicit rebalancing interventions by integrating real-world datasets, synthetic decision boundary generation, and one/few-shot scenarios. Our experimental results also reveal that the classifier's robustness is highly dependent on both data geometry and imbalance severity. Traditional models such as Decision Trees and k-NN become unreliable once the minority class falls below 25\%, ensemble methods (RF, CatBoost, LGBM) retain moderate performance down to 10\%, and only advanced models such as TabPFN, SVM, and CatBoost sustain meaningful performance at extreme imbalance (less than 5\% or one-shot settings). Our experiments further confirm that classifier performance under class imbalance is dependent on data complexity and geometry, as performance decreases when the decision boundary becomes more complicated. Our work has certain limitations, as it is based on benchmark datasets and controlled synthetic scenarios, with hyperparameters restricted to standard configurations; therefore, our results may differ on larger-scale or domain-specific applications, and further replication would strengthen generalizability. Overall, the findings highlight the intrinsic capacity of classifiers to generalize under scarcity, offering practical insights for imbalanced learning, while future directions include adaptive hyperparameter tuning, generative models for synthetic data creation, and privacy-preserving evaluation in sensitive domains.

\section*{Data Availability}

The datasets analysed during the current study are publicly available from the UCI Machine Learning Repository (\url{https://archive.ics.uci.edu/datasets}) and Kaggle Datasets (\url{https://www.kaggle.com/datasets}).  
All processed results provided in the Supplementary File and experimental scripts and code are available from the corresponding author upon reasonable request. 

\section*{Acknowledgment}
The authors acknowledge financial support from the United Arab Emirates University (Grant No. 12R313) through the Big Data Analytics Center, UAEU.

\section*{Funding Declaration}
The research is funded by the United Arab Emirates University (Grant No. 12R313) through the Big Data Analytics Center, UAEU.

\section*{Declaration of Interest}
The authors declare that they have no known competing interests.

\end{document}